\documentclass{ieeeaccess}

\newcommand\copyrighttext{%
  \footnotesize \textcopyright \the\year{} IEEE. Personal use of this material is permitted. Permission from IEEE must be obtained for all other uses, including reprinting/republishing this material for advertising or promotional purposes, collecting new collected works for resale or redistribution to servers or lists, or reuse of any copyrighted component of this work in other works.}

\newcommand\copyrightnotice{%
\begin{tikzpicture}[remember picture,overlay]
\node[anchor=south,yshift=10pt] at (current page.south) {\fbox{\parbox{\dimexpr0.75\textwidth-\fboxsep-\fboxrule\relax}{\copyrighttext}}};
\end{tikzpicture}%
}

\usepackage{cite}
\usepackage{amsmath,amssymb,amsfonts}
\usepackage{algorithmic}
\usepackage{graphicx}
\usepackage{textcomp}
\usepackage[T1]{fontenc}
\usepackage[utf8]{inputenc}
\usepackage{url}
\usepackage{booktabs} 
\usepackage{microtype}
\usepackage{inconsolata}
\usepackage{microtype}
\usepackage{adjustbox}
\usepackage{multirow}
\usepackage{makecell}
\usepackage{graphics}
\usepackage{graphicx}% http://ctan.org/pkg/graphicx
\usepackage{pifont}
\usepackage{amsfonts}
\usepackage{mathrsfs}
\usepackage[numbers]{natbib}
\usepackage{xcolor}
\usepackage{pict2e}
\usepackage{arydshln}

\newsavebox{\ORCIDlogo}
\savebox{\ORCIDlogo}{
\setlength{\unitlength}{\dimexpr 1em/256\relax}%
\begin{picture}(256,256)%
  \color[HTML]{A6CE39}\put(128,128){\circle*{256}}%
  \color{white}%
  \put(78.6,199.2){\circle*{20}}%
  \moveto(70.9,176,9)\lineto(86.3,176,9)\lineto(86.3,69.8)\lineto(70.9,69.8)%
  \closepath\fillpath%
  \moveto(108.9,176.9)\lineto(150.5,176.9)%
  \curveto(190.1,176.9)(207.5,148.6)(207.5 ,123.3)%
  \curveto(207.5,95,8)(186,69.7)(150.7,69.7)%
  \lineto(108.9,69.7)%
  \closepath\fillpath%
  \color[HTML]{A6CE39}%
  \moveto(124.3,83.6)\lineto(148.8,83.6)%
  \curveto(183.7,83.6)(191.7,110.1)(191.7,123.3)%
  \curveto(191.7,144.8)(178,163)(148,163)%
  \lineto(124.3,163)%
  \closepath\fillpath%
\end{picture}%
}

\usepackage{tikz}
\NewSpotColorSpace{PANTONE}
\AddSpotColor{PANTONE} {PANTONE3015C} {PANTONE\SpotSpace 3015\SpotSpace C} {1 0.3 0 0.2}
\SetPageColorSpace{PANTONE}%

\newcommand\orcidicon[1]{\href{https://orcid.org/#1}{\usebox{\ORCIDlogo}}}

\usepackage{hyperref} %<--- Load after everything else
\usepackage{bm}
\makeatletter
\AtBeginDocument{\DeclareMathVersion{bold}
\SetSymbolFont{operators}{bold}{T1}{times}{b}{n}
\SetSymbolFont{NewLetters}{bold}{T1}{times}{b}{it}
\SetMathAlphabet{\mathrm}{bold}{T1}{times}{b}{n}
\SetMathAlphabet{\mathit}{bold}{T1}{times}{b}{it}
\SetMathAlphabet{\mathbf}{bold}{T1}{times}{b}{n}
\SetMathAlphabet{\mathtt}{bold}{OT1}{pcr}{b}{n}
\SetSymbolFont{symbols}{bold}{OMS}{cmsy}{b}{n}
\renewcommand\boldmath{\@nomath\boldmath\mathversion{bold}}}
\makeatother

\def\BibTeX{{\rm B\kern-.05em{\sc i\kern-.025em b}\kern-.08em
    T\kern-.1667em\lower.7ex\hbox{E}\kern-.125emX}}

\begin{document}
\history{Date current version July 1, 2025.}
\doi{10.1109/ACCESS.2024.0429000}

\title{SMCLM: Semantically Meaningful Causal Language Modeling for Autoregressive Paraphrase Generation}
\author{
\uppercase{MichaŁ PereŁkiewicz}\orcidicon{0000-0001-8646-3345},
\uppercase{SŁawomir Dadas}, \uppercase{and RafaŁ PoŚwiata}}

\address{National Information Processing Institute, 00-608 Warsaw, Poland}

\markboth
{Michał Perełkiewicz \headeretal: SMCLM for Autoregressive Paraphrase Generation}
{Michał Perełkiewicz \headeretal: SMCLM for Autoregressive Paraphrase Generation}

\corresp{Corresponding author: Michał Perełkiewicz (e-mail: mperelkiewicz@opi.org.pl).}

\begin{abstract}
This article introduces semantically meaningful causal language modeling (SMCLM), a self-supervised method of training autoregressive models to generate semantically equivalent text. Our approach involves using semantically meaningful text representation as an initial embedding in the autoregressive training and generation processes. The extensive empirical study demonstrates that the SMCLM approach makes autoregressive models capable of learning robust and high-quality paraphrase generation. The proposed method is competitive with the supervised method and achieves state-of-the-art results in unsupervised approaches. This article also presents a comprehensive set of automatic metrics that cover a wide range of autogenerated paraphrase evaluation aspects. Simultaneously, this article highlights the low reliability of the metrics that are widely used in paraphrase generation evaluation, including BLEU, ROUGE, and BERTScore.
\end{abstract}


\titlepgskip=-21pt

\maketitle

\copyrightnotice

\section{Introduction}
\label{sec:introduction}
Paraphrasing can be understood as a subtle compromise between capturing the semantic meaning of original text and presenting it in alternative wording. This requires deep understanding of the nuanced meanings that are embedded in language. Despite advancements in machine learning and natural language processing (NLP), the achievement of accurate and contextually appropriate paraphrases continues to be a complex endeavor. Alongside these difficulties, automatic paraphrase generation (APG) can benefit various downstream NLP applications, such as text summarization \cite{Syed_2021}, semantic parsing \cite{wu-etal-2021-paraphrasing}, and machine translation \cite{Baranckov2014MachineTW}, and is also utilized successfully in textual data augmentation \cite{okur-etal-2022-data} and chatbot dialog systems \cite{9898139}. 

In recent years, transformer-based text generation models have achieved remarkable success in a wide range of NLP tasks, as well as producing impressive results in APG. However, such models rely primarily on a supervised learning fashion, which requires a sizable number of sentence-aligned paraphrase examples. They are also limited by the quality and coverage of the training data and the generation of domain-shifted paraphrases.
Unsupervised methods for paraphrase generation use techniques such as variational autoencoding, back-translation, or noised/ reordered text reconstruction, and do not require manually aligned sentences to generate paraphrases. However, these methods often generate text that lacks diversity and is too similar to the original text, which can result in limited expressiveness and low fluency \cite{zhou-bhat-2021-paraphrase, madnani-dorr-2010-generating}.

To address these issues, we present semantically meaningful causal language modeling (SMCLM)---a simple but effective autoregressive language model training procedure designed for semantically similar text generation. We hypothesize that injecting semantic information into initial embedding enables autoregressive language models to learn different syntax and grammar structures for semantically equivalent sentences by training with a standard self-supervised causal language modeling objective. We have proven that our technique is effective in the training of paraphrasing models that generate high-quality paraphrases. Our model can be used successfully with pretrained text-generation language models like GPT-2 and other autoregressive models. To the best of our knowledge, our method is the first that assimilates semantic meaningful embeddings and autoregressive language models to generate semantically similar text. 

We place great emphasis on reliable evaluation of our method. In addition to commonly used generated text metrics like BLEU or RougeL, we present a wide set of measures centered around following features: the quality of the generated text in terms of fluency and syntactic diversity with the input sentence and self-diversity between the generated paraphrases; semantic similarity and adequacy while penalizing surface-form similarity of the input sentence and set of and referential paraphrases. 
To ensure a fair and reliable comparison with previous methods, we reproduced the results on common paraphrase datasets for each reference method and compared them to the SMCLM results.

This article's contributions are fourfold:
\begin{enumerate}
    \item it introduces a novel approach to the training of autoregressive, semantically similar text generation models that do not require sentence-aligned data;
    \item it proposes a comprehensive set of metrics that cover a wide range of autogenerated paraphrase evaluation aspects.
    \item through extensive experiments, it proves that our method outperforms state-of-the-art unsupervised paraphrasing methods and is comparable to supervised methods (we reproduced these methods to provide a robust comparative analysis). \item its authors have released the model source code and the code of the experiments to the public\footnote{\url{https://github.com/mmichall/bleu-macaw}}, along with the pretrained general-purpose GPT2-based paraphrase model\footnote{\url{https://huggingface.co/mmichall/SMCLM-GPT-2}} and a novel, large corpus of unique English-language sentences\footnote{\url{https://huggingface.co/datasets/mmichall/smclm-10M-sentence-corpus}}.
\end{enumerate}

The remainder of this article is structured as follows: in Section \ref{sec:related_work}, we review related work about automatic semantic sentence embeddings and paraphrase generation; in Section \ref{sec:methodology}, we present an in-depth explanation of the proposed SMCLM method; Section \ref{sec:experimental_setup} describes  our experiments, used dataset and metrics; in Section \ref{sec:results}, we summarize and analyse the results and findings of our study; Sections \ref{sec:limitations} and \ref{sec:ethical_considerations} describe the limitations and ethical considerations of our study, respectively. Finally, Section \ref{sec:conclusions} concludes the paper and outlines directions for future work.

\section{Related work}
\label{sec:related_work}

\begin{table*}[!ht]
\centering
\label{tab:related_work_summary}
\def\arraystretch{1.3}
\begin{tabular}{|p{3.8cm}|p{4.5cm}|p{5.5cm}|p{2.8cm}|}
\hline
\textbf{Category} & \textbf{Technique / Approach} & \textbf{Examples / Models} & \textbf{References} \\
\hline
\multirow{5}{*}{Paraphrase Generation} 
& Rule-based / Statistical MT & Paraphrase rules, Thesaurus-based, SMT & \cite{mckeown-1983-paraphrasing, lin_pantel_2001, barzilay-lee-2003-learning, koehn-etal-2007-moses} \\
\cline{2-4}
& Encoder-Decoder (LSTM/Transformer) & LSTM-VAE, DiPS, HRQ-VAE, SynPG & \cite{prakash-etal-2016-neural, Gupta_2017, dips2019, hosking-etal-2022-hierarchical, Huang2021synpg} \\
\cline{2-4}
& Syntactic Control & POS-tagging, SOW-REAP, SynPG & \cite{Iyyer2018, goyal-durrett-2020-neural, Huang2021synpg} \\
\cline{2-4}
& Reinforcement / Adversarial / Sampling & RL, GAN, CGMH, UPSA & \cite{li-etal-2018-paraphrase, yang-etal-2019-end, cgmh, UPSA} \\
\cline{2-4}
& Pretrained Transformers (PLMs) & BART, T5, GPT-2, GPT-3 & \cite{lewis-etal-2020-bart, raffel2023exploring, Radford2019LanguageMA, niu-etal-2021-unsupervised} \\
\hline
\multirow{5}{*}{Semantic Sentence Embeddings} 
& Unsupervised Sentence Models & Skip-Thought, Sent2Vec & \cite{Kiros_2015, pgj2017unsup} \\
\cline{2-4}
& Supervised Embeddings & InferSent, USE, SBERT & \cite{conneau-etal-2017-supervised, cer2018universal, Reimers2019SentenceBERTSE} \\
\cline{2-4}
& Contrastive Learning & SimCSE, TSDAE, GTE, E5 & \cite{gao-etal-2021-simcse, DBLP:conf/emnlp/Wang0G21, li2023general, wang2022text} \\
\cline{2-4}
& Multilingual Embeddings & LaBSE, mBERT, BGE-M3 & \cite{feng2020languageagnostic, reimers-2020-multilingual-sentence-bert, chen2024bgem3embeddingmultilingualmultifunctionality} \\
\cline{2-4}
& Long-context / Autoregressive Encoding & Jina-v2, Text-embedding-3, GPT encoding & \cite{günther2024jinaembeddings28192token, Liu_Zhu_2025, Trinquier2021} \\
\hline
\end{tabular}
\caption{Summary of related work in paraphrase generation and semantic sentence embeddings}
\label{tab:related}
\end{table*}

\subsection{Paraphrase Generation} Early works on paraphrase generation focused chiefly on handcrafted \cite{mckeown-1983-paraphrasing} or automatically collected paraphrase rules \cite{lin_pantel_2001, barzilay-lee-2003-learning}, thesaurus-based approaches \cite{10.1007/978-3-540-27779-8_27, kauchak-barzilay-2006-paraphrasing}, and statistical machine-translation-based approaches \cite{wubben-etal-2010-paraphrase, koehn-etal-2007-moses}. Despite its simplicity and efficiency, the effectiveness of this method is constrained significantly by the variety of paraphrases it produces \cite{zhou-bhat-2021-paraphrase}.

In light of recent advances in deep neural network architectures, the use of deep neural models for paraphrase generation has been widely explored, particularly in the encoder-decoder architecture. Encoder and decoder pair of long short-term memory networks (LSTMs) was used by \citet{prakash-etal-2016-neural} to encode the semantic information of input sentences and pass them to the decoder to generate paraphrases. \citet{Bowman2016} propose an unsupervised generative modeling with LSTM-based Variational Autoencoder (VAE) that results in distributed latent representations of the sentences. By sampling from this representations distribution, the model generates paraphrase candidates with an LSTM decoder. \citet{Gupta_2017} extended the idea of generating paraphrases based on an LSTM-based VAE model by conditioning the encoder and decoder sides of VAE on the original sentence so that it can generate the given sentence’s paraphrases. A novel formulation of the encoder-decoder model learning problem in terms of monotone submodular function maximization, specifically targeted towards the task of paraphrasing to produce more diverse paraphrases, was presented in DiPS model \cite{dips2019}. \citet{hosking-etal-2022-hierarchical} introduce Hierarchical Refinement Quantized Variational Autoencoders (HRQ-VAE), a method for learning decompositions of dense encodings as a sequence of discrete latent variables that make iterative refinements of increasing granularity. They have proven that the use of this kind of hierarchical representation of input results in good-quality generated paraphrases. 

To generate more syntactically controllable paraphrases, \citet{Iyyer2018} incorporate part-of-speech tagging information to train an encoder-decoder model for paraphrasing. The SOW-REAP model introduced by \citet{goyal-durrett-2020-neural} incorporates the syntactic preordering of the source sentence for controlled paraphrase generation. 
\citet{Huang2021synpg} propose the syntactically controlled paraphrase generator (SynPG) model: an encoder-decoder-based model that learns to differ the semantics and the syntax of sentences from a text corpus. Other approaches to the problem of automatic paraphrase generation like reinforcement learning \cite{li-etal-2018-paraphrase, qian-etal-2019-exploring, Garg2021UnsupervisedCP}, generative adversarial networks \cite{yang-etal-2019-end, cao-wan-2020-divgan}, back-translation \cite{wieting-17-backtrans} , the Metropolis-Hastings algorithm \cite{cgmh}, and simulated annealing \cite{UPSA} have also been explored. A novel formulation of the encoder-decoder model learning problem in terms of monotone submodular function maximization, specifically targeted toward the task of paraphrasing to produce more diverse paraphrases, is presented in the DiPS model \cite{dips2019}. 

Due to the Transformer’s improved ability to capture long-term dependencies in sentences \cite{DBLP:journals/corr/VaswaniSPUJGKP17}, the use of Transformer-based models has yielded satisfactory results in automatic paraphrase generation. \citet{wang2018task} utilized a Transformer as a multi-encoder to encode not only the semantics of the input sentence but the role and frame of individual tokens, as well. \citet{roy-grangier-2019-unsupervised} present a model that combines a Vector-Quantizated AutoEncoder (VQ-VAE) \cite{10.5555/3295222.3295378} model with the Transformer network. 

Transformer-based large pretrained language models have been studied extensively in APG, including the encoder-decoder BART model \cite{lewis-etal-2020-bart}, which was pretrained on a multitask mixture of unsupervised and supervised tasks T5 model \cite{raffel2023exploring, bandel-etal-2022-quality, liu-etal-2022-paramac, PALIVELA2021100025}. The recent development of generative pretrained transformer (GPT) models \cite{Radford2019LanguageMA, Radford2018ImprovingLU, NEURIPS2020_1457c0d6}, a class of large-scale, autoregressive deep learning model, has delivered remarkable benefits in the generation of human-like text.
In APG, GPT-based models have also achieved impressive results \cite{Radford2019LanguageMA, witteveen-andrews-2019-paraphrasing, niu-etal-2021-unsupervised}.

\subsection{Semantic Sentence Embeddings}
Semantic sentence embeddings refers to a numeric representation of a sentence as a vector of real numbers, which encodes meaningful semantic information and constitutes a cornerstone in many NLP tasks. The Skip-Thought Vector unsupervised approach, introduced by \citet{Kiros_2015}, extends the neighbor prediction principles of the Word2Vec Skip-Gram model \cite{Mikolov2013EfficientEO} from individual words to entire sentences. 
The technique involves training an encoder to handle input sentences, optimizing the resulting latent representation for predicting adjacent sentences through the decoder.
Another concept of the extension of the word contexts from the Word2Vec to a larger sentence context was presented in the Sent2Vec model \cite{pgj2017unsup} with the sentence words being specifically optimized towards additive combination over the sentence, by means of the unsupervised objective function.
The InferSent model \cite{conneau-etal-2017-supervised} adopts a bi-directional LSTM network with a max-pooling operator as sentence encoder, and it is trained on the SNLI corpus \cite{conneau-etal-2017-supervised} in a supervised fashion. Another approach trained on the SNLI is the Universal Sentence Encoder (USE) \cite{cer2018universal}, released in two variants: one built on the Transformer and the other on Deep Averaging Network (DAN) \cite{iyyer-etal-2015-deep} architecture. The encoding model is designed to be more general by using multi-task learning.
The state-of-the-art Sentence-BERT (SBERT) \cite{Reimers2019SentenceBERTSE} semantically meaningful sentence embeddings based on the siamese neural network architecture composed of siamese and triplet Transformer models with tied parameters. SBERT creates sentence embeddings for a given sentence that can be compared using a cosine-similarity. \citet{9945218} presents a modified architecture of SBERT with an additional LSTM pooling layer and automatic extraction of paraphrase pairs for a target language utilizing sentence-aligned cross-lingual corpora.
Notable examples of unsupervised sentence embedding learning are methods based on denoising autoencoders like TSDAE \cite{DBLP:conf/emnlp/Wang0G21} and methods using the dropout to get slightly different encodings from the sentence embedding encoder for the same sentence in the SimCSE model \cite{gao-etal-2021-simcse}. During training, the distance between these two embeddings will be minimized, while the distance to other embeddings of the other sentences in the same batch will be maximized.
Many models have employed contrastive learning to fine-tune the BERT-based model to learn semantically meaningfull sentence representations. Weakly supervised multi-stage training was used in the GTE model \cite{li2023general} and the E5 model \cite{wang2022text}. Some contrastive learned models show comparable performance to supervised sentence embedding models and outperform the state-of-the-art SBERT \cite{10.1145/3593590} on the embedding benchmarks like MTEB \cite{muennighoff-etal-2023-mteb}.

Multilingual Sentence Transformers, such as LaBSE \citep{feng2020languageagnostic}, BGE-M3 \citep{chen2024bgem3embeddingmultilingualmultifunctionality}, and mBERT \citep{reimers-2020-multilingual-sentence-bert}, facilitate the generation of semantically comparable sentence embeddings across multiple languages. These models have been successfully employed in a variety of specialized cross-lingual tasks, including the detection of abusive clauses in Polish B2C contracts \citep{dadas_kozlowski_2024} and word alignment for non-English language pairs \citep{Wang2023MultilingualST}.

Advancements in long-context processing technologies have further enhanced the applicability of sentence embedding models to extended textual inputs. Recent models such as Text-embedding-3-large, Jina-embeddings-v2 \citep{günther2024jinaembeddings28192token}, and BGE-M3 \citep{chen2024bgem3embeddingmultilingualmultifunctionality} are capable of encoding significantly longer sequences, including full chapters or documents. This capability markedly improves the practicality and versatility of semantic vector models in real-world scenarios \citep{Liu_Zhu_2025}.

Furthermore, recent studies on the computational efficiency of autoregressive models substantiate the use of Transformer-based autoregressive architectures as a resource-efficient alternative. Although originally designed for generative purposes, these models have demonstrated competitive performance in sentence encoding tasks when properly adapted, offering notable advantages in scalability and efficiency for large-scale semantic representation learning \citep{lin-etal-2021-limitations, Trinquier2021, Shen_Song_Zhou_Chen_Liu_Zhang_Rossi_Tan_Yu_Chen_Zhou_Sun_Zhao_Wang_Gu_2025}. Table \ref{tab:related} contains summary of related work.

\section{Methodology}
\label{sec:methodology}

\begin{figure*}[!th]
\centering
\includegraphics[width=1.015\textwidth]{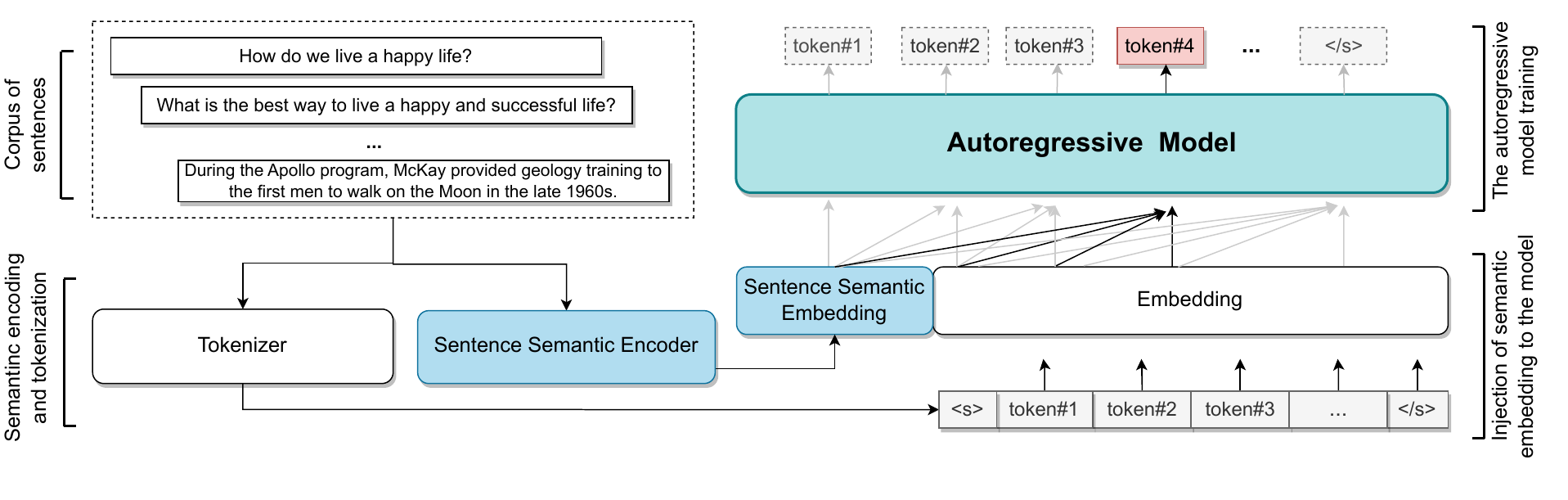}
\caption{An overview of the SMCLM learning process with an autoregressive model. First a sentence from the sentence corpus is tokenized and encoded by the sentence semantic encoder. The tokens are then embedded using the model embedding method. The semantic embedding of the whole sentence is passed, along with the embedded tokens, to the autoregressive model as an initial embedding, and is utilized in a self-supervised training process.}
\label{fig:scheme}
\end{figure*}

This section introduces the idea behind autoregression language models. It then offers a thorough explanation of the SMCLM training procedure in the context of autoregressive paraphrase generation training.

\subsection{Autoregressive language models}
An autoregressive language model is a type of generative language model that generates text by predicting each token based sequentially on the previous tokens.
Causal language modeling (CLM), a standard autoregressive model training technique, is a self-supervised paradigm in which a model is trained to predict the next token in a sequence, considering only the preceding tokens, without using any labeled data \cite{9462394}. 

Formally, for a given sequence of tokens $s_i=\{x_1,x_2, ...,x_{T_i}\}$, where $x_t \in \mathcal{V}$ for $t = 1,2, ...,T_i$, and a set of sentences $\mathcal{S} = \{s_1, s_2, ..., s_N\}$ for $i = 1,2, ...,N$ where $|\mathcal{S}| = N$, an autoregressive model learns a probability distribution $P(x_t|x_{<t})$ over the model's vocabulary $\mathcal{V}$, which is conditioned on the $x_t$ preceding tokens $x_{<t}=x_{1:t-1}=\{x_1,x_2,...,x_{t-1}\}$. $T_i$ is the number of tokens in the sequence $s_i$ and $|\mathcal{V}| = M$. A token embedding function $g : \mathcal{V} \rightarrow \mathbb{R}^{d}$ maps each token $x_j \in \mathcal{V}$ to $d$-dimensional vector $\vec{z_j} \in \mathbb{R}^{d}$ for $j = 1, 2, ..., M$. Given a model with parameters $\Theta$, the conditional probability estimation $P$ of token $x_t$ from the given sequence $s_i$ can be defined as:
\begin{equation}
    P(x_t|x_{<t};\Theta) = f(\vec{z}_{<t}; \Theta) \enspace \forall_t=1,2,...,T_i ,
    \label{eq:px_t}
\end{equation}
where $f(\vec{z}_{<t}; \Theta)$ is the function parameterized by $\Theta$ that maps the context tokens $x_{<t}$ vector representations $\vec{z}_{<t}$ to the estimated probability distribution $P$ over the vocabulary $\mathcal{V}$; $\vec{z}_{<t} = \{\vec{z}_1, \vec{z}_2,..., \vec{z}_{t-1}\} = \{g(x_1), g(x_2),..., g(x_{t-1})\}$ for $t = 1, 2, ..., T_i$. 

%$$new$$
The probability distribution $P_{s_i}$ of a token sequence $s_i$ can be defined as the product of conditional next token distributions:

%$$new$$
\begin{equation}
    P_{s_i}(x_{1:T};\Theta) = \prod_{t=1}^T P(x_t|x_{<t};\Theta)  ,
    \label{eq:px_t}
\end{equation}

%$$new$$
The autoregressive training with self-supervised CLM objective is to minimize the negative log of the probability distribution function $\ell$ defined as:

%$$new$$
\begin{equation}
    \ell(\Theta;\mathcal{S}) = - \sum_{s \in \mathcal{S}} \sum_{t=1}^T \log P(x_t|x_{<t}; \Theta),
\end{equation}

%$$new$$
by iteratively update the model parameters $\Theta$ over the set $\mathcal{S}$.

%$$new$$
% Auto-regressive language models learn a probability distribution $P(x_t|x_{<t})$ over the vocabulary $\mathcal{X}$, conditioned on the context tokens $x_{<t}=\{x_1,x_2,...,x_{t-1}\}$, where $x_t \in \mathcal{X}$ for $t = 1,2, ...,T$. 

%$$new$$
Auto-regression in language models assumes predicting the next word in a sequence based on the words that have come before it. That new sequence becomes the input to the model in its next prediction step. Auto-regressive models like recurrence neural networks and GPT are tipically trained with the standard Causal language Modeling (CLM) objective. In the CLM training, the model determines the succeeding token probability in a sequence, considering only the previous tokens in a self-supervised fashion. The process of a CLM is centred around the idea of generating text that is similar to the input text by self-supervised prediction of each token in it.

The function $f(\vec{z}_{<t}; \Theta)$ is commonly realized through the utilization of deep neural networks, such as transformers or recurrent neural networks. Such models are crafted specifically to comprehend intricate relationships among tokens, and are capable of representing a wide range of probability distributions over the vocabulary $\mathcal{V}$ \cite{math11112451}.

\subsection{Semantically meaningful CLM}
The main goal of our SMCLM training procedure is to leverage CLM objective training as semantically similar text generation training by injecting the sentence semantically meaningful representation as an initial token representation in the input sequence. For a sentence representation function $h : \mathcal{S} \rightarrow \mathbb{R}^{d}$ , we assume that if two sentences $s_1, s_2 \in \mathcal{S}$ are similar in meaning or content, their projections $h(s_1), h(s_2) \in \mathbb{R}^{d}$ are also similar in their sentence representation space of meaning\footnote{In further experiments, we define the similarity in sentence representations space as a cosine similarity.}.

During SMCLM training, for every sentence $s_i \in \mathcal{S}$, a sentence semantically meaningful representation $h(s_i) \in \mathbb{R}^d$ is prepended to the $s_i$ sentence, which results in the sequence $\widehat{s_i}=\{h(s_i),x_1,x_2, ...,x_T\}$. Therefore, the probability estimation $P'$ for the token $x_t$ is not only conditioned on the preceding tokens $x_{<t}$ in the sequence $s_i$ like in Equation \ref{eq:px_t}, but also on the semantic sentence $s_i$ representation $h(s_i)$:
\begin{equation}
    P'(x_t|\{h(s)\} \cup x_{<t};\Theta) = f(\{h(s)\} \cup z_{<t}; \Theta). %\enspace \forall_t=1,2,...,T ,
    \label{eq:px_t_smclm}
\end{equation}

Figure \ref{fig:scheme} illustrates the SMCLM training procedure. SMCLM requires a corpus of text. We used a corpus of sentences in our experiments. The SMCLM method can also be applied to longer-form text, such as paragraphs or multisentence passages, provided that semantic encoders designed for longer-form text are used.

In the SMCLM sentence paraphraser training procedure, a sentence provided from the training corpus is encoded with the encoder, obtaining sentence semantically meaningful embeddings that realize the function $h$. The given training sentence is also tokenized using a model-dependent tokenizer. Next the sentence representation is passed as initial embedding in the CLM training process, replacing the embedding of the first token---typically the special token that denotes the beginning of a sentence. As the training objective is to predict succeeding tokens based on previous ones, the model determines the probability of the following tokens from lexically and grammatically different sentences using the semantic embeddings, which are similar for semantically similar sentences. The intuition behind the SMCLM approach is that the method facilitates the memorization and generalization of diverse syntactic relationships in semantically analogous sentences during CLM training.

\section{Experimental setup}
\label{sec:experimental_setup}
Our empirical study included experiments that used our method to train a sentence paraphrasing model. We utilized the pretrained checkpoints of the GPT-2 model as the autoregressive model to be trained using SMCLM. As the semantic sentence encoder, we experimented with the pretrained bi-encoders provided in the SentenceTransformers library. We describe our GPT-2-based SMCLM implementation in detail in Appendix~\ref{sec:smclmgpt2}. 

The \textbf{GPT-2} model introduced by \citet{Radford2019LanguageMA} is a transformer-based autoregressive language model that originally contained 1.5 billion parameters. The Hugging Face library provides a handful of pretrained checkpoints of the GPT-2 model\footnote{\url{https://huggingface.co/gpt2}}. For the purpose of efficiency and to ensure that our model remained as close as possible in terms of its number of parameters with the models to which we compared it, we utilized the \textit{small} checkpoint with 768-dimensional hidden layers, 12 transform blocks, 12 attention heads, and 124 million parameters.

The \textbf{SentenceTransformers} library provides pretrained bi-encoders based on Sentence-BERT architecture\footnote{\url{https://www.sbert.net/}} \cite{reimers-2020-multilingual-sentence-bert}. The bi-encoders compute semantically meaningful text representations for more than 100 languages. They are trained on a wide range of datasets to ensure robust embeddings. Our SMCLM implementation utilizes the \textbf{paraphrase-mpnet-base-v2}\footnote{\url{https://huggingface.co/sentence-transformers/paraphrase-mpnet-base-v2}}; the MPNet-based encoder offers a high quality of semantic embeddings\footnote{We also experimented with \textbf{all-mpnet-base-v2} and \textbf{all-roberta-large-v1} models.}. 

\subsection{Datasets}
\label{sec:datasets}
We utilized three datasets of paraphrases for evaluation: Quora Question Pairs (QQP), Microsoft Common Objects in Context (MSCOCO), and CNN/Daily Mail News (CNN News). 
\paragraph{QQP}\footnote{\url{https://www.kaggle.com/c/quora-question-pairs}} This dataset contains short questions derived from Quora\footnote{\url{https://www.quora.com/}}---a community-based question-and-answer website. The dataset comprises over 400 000 question pairs, each annotated with a binary value that indicates whether two questions are paraphrasal. More than 140 000 of them have been annotated as correct paraphrases, and we limited our further experiments exclusively to this group.
\paragraph{MSCOCO} 
This dataset comprises a popular collection of human-annotated captions for over 120 000 images, each accompanied by five captions provided by distinct annotators, recognized as a standard benchmark dataset for the image caption generation task\footnote{\url{https://cocodataset.org/}} \cite{10.1007/978-3-319-10602-1_48}. Most captions focus on describing the most relevant object or action in an image. We employed the dataset split proposed by Karpathy and Fei-Fei\footnote{\url{https://huggingface.co/datasets/phiyodr/coco2017}} \cite{Karpathy2014DeepVA}.
\paragraph{CNN News}
This dataset contains over 80 000 groups of paraphrases created using ChatGPT by the generation of five paraphrases for questions and sentences derived from the CNN News dataset\footnote{\url{https://github.com/abisee/cnn-dailymail}, \url{https://huggingface.co/datasets/humarin/chatgpt-paraphrases}} \cite{see-etal-2017-get}. This dataset is characterized by longer passages of text than those found in the others.

In the cases of the QQP and CNN News datasets, in which official data splits are unavailable, we partitioned the data into training, validation, and test sets at a ratio of 80\%, 5\%, and 15\%, respectively. To prevent repetition of sentences across dataset splits, we organized the sentences into paraphrase groups in each of the datasets and assigned an entire paraphrase group to a particular dataset split. We then randomly selected one sentence from each paraphrase group and created sentence--paraphrase pairs for the selected sentence and correlating paraphrases, creating examples for supervised learning. 

In addition to the datasets for supervised learning, we prepared corpora for unsupervised learning. We unsqueezed the splits into simple text corpora that comprise per-line sentences. Thus we obtained training and validation corpora for unsupervised learning methods. The test data for both supervised and unsupervised method evaluations included sentence--list pairs of reference paraphrases\footnote{Prepared splits are available at \url{https://huggingface.co/datasets/mmichall/smclm-qqp}, \url{https://huggingface.co/datasets/mmichall/smclm-cnn-news} and \url{https://huggingface.co/datasets/mmichall/smclm-mscoco}}. Table \ref{tab:datasets} presents the numbers of examples in the data splits.

\paragraph{10M Sentence Corpus}
The pretrained GPT-2 model required fine-tuning on a corpus of English sentences to train the relationship between initial semantic embeddings and training examples in our SMCLM approach. In addition to the datasets described above, we created the novel 10M Sentence Corpus: a corpus of ten million unique English sentences. We used it to train the general-purpose SMCLM model in our experiments. The 10M Sentence Corpus is described in detail in Appendix \ref{app:10m}. 
\begin{table}[]
\small
\setlength{\tabcolsep}{6pt} % Default value: 6pt
\renewcommand{\arraystretch}{1.1} % Default value: 1
\begin{tabular}{lcccccc}
\toprule
& \multicolumn{2}{c}{Supervised} & & \multicolumn{2}{c}{Unsupervised} & \\
\cmidrule(rl){2-3} \cmidrule(rl){5-6}
Dataset & Train & Valid & & Train & Valid & Test \\ \hline
QQP    &  120k&  7k& &  188k&  12k&  13k\\
MSCOCO   &  500k&  24k& &  591k&  30k&  5k\\
CNN News &  320k&  20k& &  384k&   24k&  12k\\
\bottomrule
\end{tabular}
\caption{The approximate sizes of the data splits for each dataset under study.}
\label{tab:datasets}
\end{table}

\subsection{Models for comparison}

The selection of the models with which we compared our solution was dictated by their performance described in publications, the year of publication (we preferred newer ones), and the availability of the source code (due to our reproduction of each method). We considered the following models:
\begin{enumerate}
\item \textbf{CopyInput} -- a baseline that copies the source sentence as the output directly, without paraphrasing; 
\item \textbf{GPT 3.5 and GPT-4o-mini} -- we generated five paraphrases of the source sentence by requesting the ChatGPT API\footnote{\url{https://openai.com/}} with the prompt "Present five paraphrases of the following sentence: \textit{source sentence}". We employed the gpt-3.5-turbo-1106 and gpt-4o-mini-2024-07-18 models\footnote{\url{https://platform.openai.com/docs/models}}. Due to the large size of these models behind ChatGPT, we used them as a reference methods. In the remainder of this work, we refer to these methods collectively as \textbf{ChatGPT}; 
\item \textbf{DiPS}: a sequence-to-sequence model that maximizes a submodular objective function specifically targeted toward paraphrasing\footnote{\url{https://github.com/malllabiisc/DiPS}} \cite{dips2019}; 
\item \textbf{QCPG} -- a T5-based encoder-decoder model trained on the task of controlled paraphrase generation\footnote{\url{https://github.com/IBM/quality-controlled-paraphrase-generation}} \cite{bandel-etal-2022-quality}; 
\item \textbf{GPT-2} -- a fine-tuned GPT-2 checkpoint with which we used prepared paraphrase pairs for sequence-to-sequence model fine-tuning; 
\item \textbf{T5 v1.1} -- Google's T5v1.1 model version with approximately 250 million parameters, fine-tuned using our datasets \footnote{\url{https://huggingface.co/google/t5-v1\_1-base}} \cite{raffel2023exploring}. We trained the model in a sequence-to-sequence manner; 
\item \textbf{BackTransl} -- a back-translation method that translates an English sentence to five target languages: French, German, Spanish, Italian, and Hungarian. We translated the results back into English, generating five paraphrase candidates\footnote{We utilize pre-trained OPUS-MT translation models \cite{TiedemannThottingal:EAMT2020} publicly available by the University of Helsinki (\url{https://opus.nlpl.eu/Opus-MT}).}; 
\item \textbf{CorrputLM} -- a GPT-2 model trained in an unsupervised manner to reconstruct corrupted input sentences \cite{DBLP:journals/corr/abs-2006-05477}; 
\item \textbf{CGMH} -- an unsupervised approach that uses Metropolis-Hastings sampling for constrained sentence generation\footnote{\url{https://github.com/NingMiao/CGMH}.} \cite{cgmh}; 
\item \textbf{SMCLM} -- a GPT-2 model trained using our SMCLM method. 
\end{enumerate}

We recorded the results for the model trained with prepared text corpora for every dataset under study (see Section \ref{sec:datasets}). Moreover, we trained models using the 10M Sentence Corpus without and with fine-tuning on the dataset corpora. In Table \ref{tab:res}, these SMCLM models are named \textbf{SMCLM}, \textbf{SMCLM-10M}, and \textbf{SMCLM-10M-ft}, respectively. 

\textbf{GPT-2}, \textbf{CorruptLM}, and all three \textbf{SMCLM} models are based on the 124-million-parameter Hugging Face GPT-2 checkpoint\footnote{\url{https://huggingface.co/gpt2}}. The \textbf{QCPG} model uses the \textit{base} Hugging Face checkpoint of a T5 model with 223 million parameters\footnote{\url{https://huggingface.co/t5-base}}. We used training hyperparameters recommended by the authors of the methods with which we compared ours. For the SMCLM, GPT-2, and T5 v1.1 methods' training, we used the hyperparameters that had achieved the best results from the ranges studied. The selection of the hyperparameters is described in Appendix \ref{app:params}. We used Hugging Face’s Transformers library to train our SMCLM models\footnote{\url{https://huggingface.co/docs/transformers}}.

\subsection{Candidate paraphrase generation}
We generated five paraphrase candidates of each test sentence for each evaluated method. From these, we selected the best one according to the SBERT-\textit{i}BLEU measure of the input sentence--candidate pair (see Section \ref{sub:eval} for details of the measure). We utilized the \textit{diverse beam search} decoding algorithm \cite{Vijayakumar2016DiverseBS} or recommended one for the method if it was described in the method's publication. For the SMCLM method, we generated paraphrase candidates by passing semantic sentence encoding to the trained model and autoregressively generating candidates using the \textit{diverse beam search} algorithm\footnote{Parameters of \textit{diverse beam search} are described in Appendix \ref{app:params}.}.

\subsection{Automatic evaluation}
\label{sub:eval}
Automatic evaluation in APG is a complex task. Widely used metrics like BLEU \cite{bleu} or ROUGE \cite{lin-2004-rouge} are based on the lexical similarity of text and depend heavily on the number of gold reference paraphrases to which they compare candidates. They have been shown to suffer from low correlation with human judgment \cite{ng-abrecht-2015-better, novikova-etal-2017-need, chen-etal-2019-evaluating}. 
The evaluation of short-sentence paraphrases also requires special attention. The BLEU metric measures the word similarity between a paraphrase candidate and the gold references using n-grams, contiguous sequences of n-words (typically four). It can yield zero similarity for short sentences that differ by a single word. Evaluation of generated paraphrases also requires high-quality measures of semantics. We present a broad set of robust evaluation metrics to address these issues. We evaluated the \textbf{lexical similarity} of a source sentence and generated paraphrases, the \textbf{lexical diversity} of generated paraphrases, and the best candidate and references. The \textbf{fluency} of generated text and \textbf{semantic similarity}, both simple and lexically dependent, are other important factors we evaluated. 
 
\textbf{The lexical similarity} of the best candidate and gold references was measured using BLEU-3 and ROUGE-L as standard metrics used in APG. Values of 1 indicate a total overlap between sentences; values of 0 indicate no common words.

With regard to  \textbf{lexical diversity}, we employed BLEU-based metrics---\textit{ori}BLEU and \textit{self}BLEU \cite{Zhu2018TexygenAB}---to assess the linguistic diversity of the generated candidates. The former gauges the lexical similarity between the source sentence and the set of candidates. The latter considers how one candidate resembles the rest in a set of generated candidates. To accommodate short sentences, we relied on a three-gram-based BLEU-3 metric. Regarding one candidate as a hypothesis and the others as a reference, we could calculate the BLEU-3 score for every generated candidate. We then defined the average BLEU-3 score as the \textit{self}-BLEU for all possible candidate pairs. Lower values indicate greater diversity among generated paraphrases. We used text normalization to lower-case and remove punctuation for the linguistic diversity and similarity metrics. Values close to 0 indicate high linguistic diversity between the sentences being compared.

\textbf{The fluency} of the text generated was measured using the pretrained BERT model, which had been trained to estimate the fluency and grammatical correctness of the input text as a regression score\footnote{\url{https://huggingface.co/prithivida/parrot\_fluency\_model}}. Values close to 1 indicate high text fluency.

\textbf{Semantic similarity} is measured by averaging the semantic similarities of the best generated candidate--gold reference pairs. We used the BERTScore (BERT for short) \cite{DBLP:conf/iclr/ZhangKWWA20} and SBERT metrics to ascertain the semantic similarity of the sentence pairs. We also evaluated the semantic similarity between source sentence--generated paraphrase pairs. For this purpose, we used the same measures as before, \textit{ori}BERT and \textit{ori}SBERT. We motivated source sentence--generated paraphrase comparison by measuring the semantic similarity for all generated paraphrases---not only for the best, as we had done previously. The SBERT and \textit{ori}SBERT metrics were based on pretrained bi-encoders from the SentenceTransformers library\footnote{For evaluation purposes, we use the paraphrase-distilroberta-base-v2 model due to its proven quality of paraphrase embeddings. Note that we use a different Sentence-Transformer model for evaluation that we use to train our SMCLM model of reliable assessment.}. We determined the semantic similarity as a cosine similarity between the vector representations in the pairs of sentences being compared.

\begin{table*}[!th]
\vspace{0.3cm}
\resizebox{2.1\columnwidth}{!}{
    \begin{tabular}{llcccccccccccccc}
    \toprule
    \multicolumn{2}{r}{} & \multicolumn{11}{c}{QQP} \\ \cmidrule(rl){3-13}
    \multicolumn{2}{r}{} & \multicolumn{2}{c}{lexical diversity} & \multicolumn{2}{c}{lexical similarity} & \multicolumn{1}{c}{fluency} & \multicolumn{4}{c}{semantic similarity} &  \multicolumn{2}{c}{lexically-dependent semantic sim}\\ 
    \cmidrule(rl){3-4} \cmidrule(rl){5-6} \cmidrule(rl){7-7} \cmidrule(rl){8-11} \cmidrule(rl){12-13}
    \multicolumn{1}{r}{} & Model & \textit{ori}BLEU $\downarrow$  & \textit{self}BLEU $\downarrow$ &  BLEU-3 & ROUGE-L & fluency & \textit{ori}BERT & BERT & \textit{ori}SBERT & SBERT & \textbf{BERT-\textit{i}BLEU} & \textbf{SBERT-\textit{i}BLEU} \\ \midrule
    \multirow{3}{*}{\small References} 
    & CopyInput     & 100.0 & 100.0 & 36.51 & 55.43 & 88.35 & 100.0 & 88.53 & 100.0 & 87.62 & 0.0 & 0.0    \\
    & GPT 3.5       & 36.41 & 33.70 & 7.0   & 29.47 & 94.98 & 88.04 & 81.52 & 83.59 & 79.53 & 86.05 & 87.32 \\
    & GPT-4o-mini   & 32.12 & 31.90 & 8.2   & 28.81 & 97.17 & 90.01 & 82.65 & 83.40 & 79.61 & 86.20 & 87.41 \\
    \cmidrule(rl){1-13}
    \multirow{4}{*}{\small Supervised} 
     & DiPS  & 55.87 & \textbf{37.73} & \textbf{20.33} & \textbf{39.91} & 78.14 & 86.30 & 80.87 & 55.23 & 52.82 & 74.74 & 54.65 \\ 
     & QCPG  & 43.20 & 50.81    & 9.68 & 33.42 & 88.88 & \textbf{87.34} & 80.60 & 68.91 & 72.07 & 84.01 & 80.19 \\
     & GPT-2 &  \textbf{7.60} & 	61.74 & 9.70 & 	34.16 & 87.07 & 86.35 & \textbf{82.70} & 75.50 & 76.78 & \textbf{87.67} & 86.07 \\
     & T5 v1.1 & 43.19 & 48.19 & 12.21 & 33.08 & \textbf{94.28} & 88.49 & 81.71 & \textbf{83.95} & \textbf{82.05} & 84.67 & 	\textbf{86.82} \\
   \cmidrule(rl){1-13}
    \multirow{6}{*}{\small \makecell[l]{Unsupervised}} 
     & BackTransl    & 88.27 & 		82.11 & 	\textbf{22.78} & \textbf{45.50} & 	\textbf{91.69} & \textbf{97.94} & 	\textbf{85.52} & 	\textbf{93.54} & 	\textbf{81.52} & 	71.46 & 	70.90 \\
     & CorruptLM   & \textbf{51.57} & 		\textbf{54.46} & 		12.55 & 	36.61 & 	82.60 & 	91.43 & 	82.24 & 	79.74 & 	73.61 & 	83.08 & 	78.96 \\ 
    % & CGMH\_fast   & 70.90 &    88.31 &     20.57 &     43.53 &     73.95 &     89.94 &     80.95 &     58.28 &     53.40 &  69.19 &     52.87 \\
     & CGMH             & 70.71 &   85.08 &     19.07 &     42.42 &     72.97 &     89.83 &     80.54 &     53.12 &     50.96 &     73.58 &     54.71 \\ \cdashline{3-13}
     & SMCLM & 56.91 & 		63.41 & 		16.77 & 	37.34 & 	91.18 & 	90.75 & 	83.43 & 	79.81 & 	75.74 & 	83.67 & 	80.59 \\
     & SMCLM-10M     & 54.88 & 		64.68 & 		13.67 & 	40.20 & 	85.99 & 	90.51 & 	83.37 & 	83.48 & 	79.39 & 	83.90 & 	83.98 \\
     & SMCLM-10M-ft & 61.86 & 		65.62 & 		15.54 & 	40.78 & 	87.81 & 	91.76 & 	84.19 & 	85.35 & 	80.50 & 	\textbf{84.52} & 	\textbf{84.80} \\
     \bottomrule
    \end{tabular}
}

\resizebox{2.1\columnwidth}{!}{
    \begin{tabular}{llcccccccccccc}
    \toprule
    \multicolumn{2}{r}{} & \multicolumn{11}{c}{MSCOCO} \\ \cmidrule(rl){3-13}
     \multicolumn{2}{r}{} & \multicolumn{2}{c}{lexical diversity} & \multicolumn{2}{c}{lexical similarity} & \multicolumn{1}{c}{fluency} & \multicolumn{4}{c}{semantic similarity} &  \multicolumn{2}{c}{lexically-dependent semantic sim}\\ 
    \cmidrule(rl){3-4} \cmidrule(rl){5-6} \cmidrule(rl){7-7} \cmidrule(rl){8-11} \cmidrule(rl){12-13}
    \multicolumn{1}{r}{} & Model & \textit{ori}BLEU $\downarrow$  & \textit{self}BLEU $\downarrow$ &  BLEU-3 & ROUGE-L & fluency & \textit{ori}BERT & BERT & \textit{ori}SBERT & SBERT & \textbf{BERT-\textit{i}BLEU} & \textbf{SBERT-\textit{i}BLEU} \\ \midrule
    \multirow{3}{*}{\small References} 
    & CopyInput     & 100.0 & 100.0 & 22.22 & 22.35 & 84.46 & 100.0 & 82.34 & 100.0 & 65.50 & 0.0 & 0.0   \\
    & GPT 3.5       & 52.80 & 42.43 & 9.70 & 19.26 & 93.31 & 89.13 & 80.17 & 87.22 & 63.84 & 85.05 & 86.85 \\
    & GPT-4o-mini   & 50.41 & 40.89 & 9.81 & 20.96 & 93.40 & 90.03 & 81.19 & 87.60 & 64.14 & 85.25 & 86.91 \\ 
    \cmidrule(rl){1-13}
    
    \multirow{4}{*}{\small Supervised} 
     & DiPS & 40.02 & \textbf{31.99} & 22.39 & 	21.81 & \textbf{91.03} & 84.15 & 81.52 & 67.79 & 58.18 & 79.60 & 70.04 \\ 
     & QCPG & 38.96 & 45.09 & 22.23 & \textbf{23.06} & 84.97 & 85.02 & 81.71 & 66.53 & 61.52 & 82.82 & 78.81 \\
     & GPT-2 &  \textbf{0.75} & 	59.52 & 19.37 & 22.21 & 88.45 & 81.94 & 82.83 & 65.73 & 60.48 & \textbf{85.78} & 	79.25 \\
     & T5 v1.1 & 42.72 & 55.14 & \textbf{23.13} & 21.93 & 83.82 & \textbf{86.95} & \textbf{83.07} & \textbf{75.91} & \textbf{62.89} & 84.42 & 	\textbf{81.23}   \\ 
    \cmidrule(rl){1-13}
    
    \multirow{6}{*}{\small \makecell[l]{Unsupervised}} 
     & BackTransl    & 87.89 & 		75.60 & 		13.71 & 	19.47 & 	85.55 & 	\textbf{97.32} & 	80.74 & \textbf{91.64} & 	62.46 & 	77.29 & 	76.76 \\
     & CorruptLM    & 44.0 & 		\textbf{38.83} & 		4.83 & 	15.86 & 	73.82 & 	87.0 & 	76.49 & 	68.41 & 	53.44 & 	83.68 & 	77.77   \\ 
     & CGMH & 77.60 & 89.06 & 16.11 & 20.47 & 74.71 & 91.99 & 79.47 & 69.84 & 50.34 & 73.31 & 61.43 \\ \cdashline{3-13}
     & SMCLM & 56.52 & 		73.46 & 		\textbf{21.36} & 	\textbf{22.32} & 	\textbf{87.59} & 90.52 & \textbf{82.17} & 85.91 & \textbf{62.98} & 	81.32 & 	81.20 \\
     & SMCLM-10M     & \textbf{32.71} & 		52.77 & 		11.50 & 	19.43 & 85.27 & 84.25 & 79.54 & 76.89 & 	60.75 & 	84.28 & 	84.40 \\
     & SMCLM-10M-ft & 43.33 & 		64.41 & 		17.80 & 	21.82 & 	84.47 & 	87.14 & 81.36 & 	81.22 & 	62.47 & 	\textbf{84.74} & 	\textbf{85.03}  \\

     \bottomrule
    \end{tabular}
}

\resizebox{2.1\columnwidth}{!}{
    \begin{tabular}{llcccccccccccc}
    \toprule
    \multicolumn{2}{r}{} & \multicolumn{11}{c}{CNN News} \\ \cmidrule(rl){3-13}
    \multicolumn{2}{r}{} & \multicolumn{2}{c}{lexical diversity} & \multicolumn{2}{c}{lexical similarity} & \multicolumn{1}{c}{fluency} & \multicolumn{4}{c}{semantic similarity} &  \multicolumn{2}{c}{lexically-dependent semantic sim}\\ 
    \cmidrule(rl){3-4} \cmidrule(rl){5-6} \cmidrule(rl){7-7} \cmidrule(rl){8-11} \cmidrule(rl){12-13}
    \multicolumn{1}{r}{} & Model & \textit{ori}BLEU $\downarrow$  & \textit{self}BLEU $\downarrow$ &  BLEU-3 & ROUGE-L & fluency & \textit{ori}BERT & BERT & \textit{ori}SBERT & SBERT & \textbf{BERT-\textit{i}BLEU} & \textbf{SBERT-\textit{i}BLEU} \\ \midrule
    \multirow{3}{*}{\small References} 
        & CopyInput   & 100.0 & 100.0 & 54.11 & 24.55 & 86.19 & 100.0 & 87.99 & 100.0 & 84.83 & 0.0 & 0.0  \\ 
        & GPT 3.5     & 47.99 & 49.48 & 47.33 & 23.81 & 94.34 & 86.66 & 87.80 & 82.66 & 84.30 & 82.24 & 82.61 \\ 
        & GPT-4o-mini & 45.19 & 47.11 & 47.60 & 23.80 & 95.50 & 87.98 & 88.79 & 83.16 & 85.10 & 83.14 & 83.11 \\
        \cmidrule(rl){1-13}
    \multirow{4}{*}{\small Supervised} 
     & DiPS &  47.14 & 	\textbf{64.62} & 28.63 & 18.98 & 71.74 & 79.37 & 75.30 & 62.0 & 57.99 & 72.05 & 61.41 \\
     & QCPG & 26.0 & 65.43 & 10.84 & 13.05 & 78.92 & 80.72 & 77.55 & 65.44 & 66.63 & 81.57 & 76.31 \\
     & GPT-2 &  \textbf{3.20} & 65.46 & 17.58 & 18.20 & 89.97 & 81.76 & 81.95 & 75.09 & 76.30 & \textbf{85.56} & 	\textbf{84.67}   \\
     & T5 v1.1 & 63.77 & 76.79 & \textbf{46.64} & \textbf{26.07} & 	\textbf{90.94} & \textbf{88.44} & \textbf{83.71} & \textbf{88.01} & \textbf{83.65} & 76.16 & 	79.24   \\ 
    
    \cmidrule(rl){1-13}
    \multirow{6}{*}{\small \makecell[l]{Unsupervised}} 
     & BackTransl    & 81.02 & 72.19 & \textbf{32.01} & \textbf{21.90} & 86.25 & \textbf{95.57} & \textbf{85.34} & \textbf{89.41} & 	\textbf{80.95} & 	81.64 & 	76.87  \\
     & CorruptLM    & 55.79 & 	52.99 & 		20.63 & 	18.70 & 	75.20 & 	88.40 & 	80.63 & 	75.19 & 	72.78 & 	\textbf{81.86} & 	73.83   \\ 
     & CGMH & 51.13 & 89.73 & 18.0 & 14.25 & 45.98 & 80.81 & 74.01 & 48.90 & 48.12 & 72.37 & 55.50 \\ \cdashline{3-13}
     & SMCLM & 19.19 & 	\textbf{41.62} & 9.69 & 	16.38 & \textbf{95.03} & 	79.34 & 	76.41 & 	67.43 & 	64.59 & 	79.87 & 	76.17 \\
     & SMCLM-10M  & \textbf{18.60} & 	47.73 & 8.99 & 	14.41 & 	90.81 & 	78.23 & 	76.68 & 	67.89 & 	68.27 & 	80.89 & 	78.14  \\
     & SMCLM-10M-ft & 18.64 & 		45.81 & 		9.76 & 	14.57 & 	91.84 & 	78.37 & 	77.13 & 	68.17 & 	68.71 & 	81.07 & 	\textbf{78.37}    \\
     \bottomrule 
    \end{tabular}
}
    
    \bigskip
\resizebox{2.1\columnwidth}{!}{
    \begin{tabular}{llcccccccccccc}
      \toprule
    \multicolumn{2}{r}{} & \multicolumn{11}{c}{Total} \\ \cmidrule(rl){3-13}
    \multicolumn{2}{r}{} & \multicolumn{2}{c}{lexical diversity} & \multicolumn{2}{c}{lexical similarity} & \multicolumn{1}{c}{fluency} & \multicolumn{4}{c}{semantic similarity} &  \multicolumn{2}{c}{lexically-dependent semantic sim}\\ 
    \cmidrule(rl){3-4} \cmidrule(rl){5-6} \cmidrule(rl){7-7} \cmidrule(rl){8-11} \cmidrule(rl){12-13}
    \multicolumn{1}{r}{} & Model & \textit{ori}BLEU $\downarrow$  & \textit{self}BLEU $\downarrow$ &  BLEU-3 & ROUGE-L & fluency & \textit{ori}BERT & BERT & \textit{ori}SBERT & SBERT & \textbf{BERT-\textit{i}BLEU} & \textbf{SBERT-\textit{i}BLEU} \\ \midrule
    \multirow{3}{*}{\small References} 
        & CopyInput   & 100,00 & 100,00 & 37,61 & 34,11 & 86,33 & 100,00 & 86,29 & 100,00 &	79,32 &	0,00 & 0,00  \\ 
        & GPT 3.5     & 45,73 &	41,87 &	21,34 &	24,18 &	94,21 &	87,94 &	83,16 &	84,49 &	75,89 &	84,45 &	85,59 \\ 
        & GPT-4o-mini & 42,57 &	39,97 &	21,87 &	24,52 &	95,36 &	89,34 &	84,21 &	84,72 &	76,28 &	84,86 &	85,81 \\ \cmidrule(rl){1-13}
    \multirow{4}{*}{\small Supervised} 
     & DiPS &       47,68 &	        \textbf{44,78}  &	17,12         &	26,90 &	80,30 &	83,27 &	79,23 &	61,67 &	56,33 &	75,46 &	62,03 \\
     & QCPG &       36,05 &	        53,78           &	14,25         &	23,18 &	84,26 &	84,36 &	79,95 &	66,96 &	66,74 &	82,80 &	78,44 \\
     & GPT-2 &      \textbf{3,85} &	62,49           &	15,55         &	20,21 &	88,50 &	83,35 &	82,49 &	72,11 &	71,19 &	\textbf{86,34} &	\textbf{83,33}   \\
     & T5 v1.1 &    49,89 &	        60,04           &	\textbf{27,33} &	\textbf{27,03} &	\textbf{89,68} &	\textbf{87,96} &	\textbf{82,83} &	\textbf{82,62} &	\textbf{76,20} &	81,75 &	82,43  \\ 
    
    \cmidrule(rl){1-13}
    \multirow{6}{*}{\small \makecell[l]{Unsupervised}} 
     & BackTransl    & 85,73 &	76,63 &	\textbf{22,83} & \textbf{28,96} &	87,83 &	\textbf{96,94} &	\textbf{83,87} &	\textbf{91,53} &	\textbf{74,98} &	76,80 &	74,84  \\
     & CorruptLM    & 50,45 &	\textbf{48,76} &	12,67 &	23,72 &	77,21 &	88,94 &	79,79 &	74,45 &	66,61 &	82,87 &	76,85   \\ 
     & CGMH         & 66,48 &    87,96 &	17,73 &	25,71 &	64,55 &	87,54 &	78,01 &	57,29 &	49,81 &	73,09 &	57,21 \\ \cdashline{3-13}
     & SMCLM & 44,21 &	59,50 &	15,94 &	25,35 &	\textbf{91,27} &	86,87 &	80,67 &	77,72 &	67,77 &	81,62 &	79,32 \\
     & SMCLM-10M  & \textbf{35,40} &	55,06 &	11,39 &	24,68 &	87,36 &	84,33 &	79,86 &	76,09 &	69,47 &	83,02 &	82,17  \\
     & SMCLM-10M-ft & 41,28 &	58,61 &	14,37 &	25,72 &	88,04 &	85,76 &	80,89 &	78,25 &	70,56 &	\textbf{83,44} &	\textbf{82,73}   \\
     \bottomrule 
    \end{tabular}
}

    % \caption{Paraphrase generation results on the QQP, MSCOCO, CNN News and PAWS-X datasets grouped in supervised and unsupervised sets. The results of the CopyInput and ChatGPT methods are presented as references. The dashed line denotes the results of our SMCLM method. The bolded values indicate the best results in the supervised and unsupervised groups.}
    \label{tab:res}
\caption{Paraphrase generation results on the QQP, MSCOCO, and CNN News datasets, as well as the average results from these datasets (Total), are grouped into supervised and unsupervised sets. The results of the CopyInput, GPT 3.5 and GPT-4o-mini methods are presented as references. The dashed line denotes the performance of our SMCLM method. Bolded values indicate the best results within the supervised and unsupervised groups.}
\label{tab:res}
\end{table*}

\paragraph{Lexically-dependent semantic similarity.}
Inspired by \citet{Niu2021UnsupervisedPW}, we use the BERT-\textit{i}BLEU and SBERT-\textit{i}BLEU metrics, as \textbf{lexically-dependent semantic similarity} measures. The measures promote the semantic similarity of a source sentence and a candidate, while penalizing the similarity of surface form between them:

\[
    \small {
        \textrm{(S)BERT-\textit{i}BLEU} = \left(\frac{\beta * \textrm{B}^{-1} + (1-\textrm{\textit{b}BLEU})^{-1}} {\beta+1.0} \right)^{-1}
    }
    \label{eq:sbert}
\]

where B = BERT(source, best) for BERT\textit-{i}BLEU, B = SBERT(source, best) for SBERT\textit-{i}BLEU, and \textit{b}BLEU = BLEU-3 (source, best).
Like \citet{Niu2021UnsupervisedPW}, to ensure consistency in score ranges, we applied a scaling factor $\beta$ to adjust BERT(source, best) and SBERT(source, best) scores, aligning them with the range of \textit{b}BLEU. In our experiments, we used a scaling factor of $\beta$=2 (as opposed to the original $\beta$=4), which was derived from the average difference in range between the values of BERT(source, best)/ SBERT(source, best) and \textit{b}BLEU in all paired input-reference sentences from all three datasets: QQP, MSCOCO and CNN News\footnote{A detailed analysis of the scaling factor $\beta$ in the \textrm{(S)BERT-\textit{i}BLEU} equation is presented in Appendix \ref{sec:e}}. This normalization allows for a more consistent evaluation framework. Accordingly, the computed measures enable a comparative analysis between the input and the corresponding reference paraphrases across the three datasets employed in this study. To compute BLEU, BLEU-based, ROUGE, and BERTScore metrics, we utilized Hugging Face’s Evaluate library\footnote{\url{https://huggingface.co/docs/evaluate}}.

\section{Results and analysis}
\label{sec:results}
To achieve a more robust evaluation, we reproduced the results for the methods to which we compared our solution. Table \ref{tab:res} presents the results for each of the methods.

\subsection{Lexical Diversity} Regarding lexical diversity, for most methods, \textit{ori}BLEU's results indicate low or medium varieties of generated paraphrases from the source sentences. The exception is the GPT-2 model, which achieves distinctly better results in this measure by generating more distinct passages of text. Despite this, \textit{self}BLEU indicates that the variety of paraphrases generated by the GPT-2 model is not superior to the other methods. The BackTransl method generates the candidates that are the most similar to the source sentence and with the least diversity. The diversity of the generated paraphrases was notably better using the DiPS method for short sentences and questions (the MSCOCO and QQP datasets, respectively). The SMCLM method achieved respectable results in terms of diversity, particularly for longer sentences from the CNN News dataset. The linguistic diversity of the SMCLM method was the best for the model trained on the 10M Sentence Corpus (SMCLM-10M) and slightly worse when fine-tuned (SMCLM-10M-ft) and trained on the dataset corpora (SMCLM). 

\subsection{Lexical Similarity} The lexical similarity of the best candidate to the gold truths measured by the BLEU and ROUGE-L metrics varied depending on the dataset under study. These differences can be explained by each dataset necessiting a different way of generating gold truth paraphrases. It is important to note that in datasets such as QQP and MSCOCO, the gold standard references tend to be lexically similar to the source sentences. This characteristic inherently favors models that generate low-diversity paraphrases and penalizes those producing more varied expressions when evaluated using n-gram-based metrics such as BLEU-3. We highlight this limitation early to motivate the need for evaluating semantic similarity beyond surface-level overlap. We used the BLEU-3 and ROUGE values of the CopyInput and ChatGPT methods for word-level verification of the gold truth paraphrases. For the QQP and MSCOCO datasets, the CopyInput results of BLEU-3 and ROUGE-L suggest high linguistic similarity between the source sentences and the corresponding gold references.The candidates generated by ChatGPT achieved better linguistic diversity while simultaneously achieving high semantic similarity for the best-generated candidates' metrics, such as BERT and SBERT. For the CNN News dataset, on which ChatGPT generated the gold truth, the results of lexical diversity for the CopyInput and ChatGPT methods are comparable. These results lead us to conclude that the QQP and MSCOCO gold reference selection methods promote sentences with high lexical similarity. These results confirm our hypothesis that the sensitivity of the BLEU and ROUGE methods to the quality of the references used and their lack of consideration for semantic information results in the need for further research on the metrics used to assess the quality of APG methods.

\subsection{Fluency} The fluency measure indicates that supervised methods, particularly GPT-2 and T5, generate more fluent English text. The CGMH method achieved the lowest fluency score for all datasets; however, other unsupervised methods performed better in this category, particularly BackTransl and SMCLM. Interestingly, models trained on the 10M Sentence Corpus scored slightly lower on the fluency metric. We hypothesize that this is due to their generation of more diverse lexical and syntactic structures, which may diverge from the canonical sentence patterns learned from narrower datasets. Moreover, the automated fluency model itself may exhibit a bias toward more typical or frequent constructions, potentially penalizing creative or less conventional — yet still fluent — paraphrases.

\subsection{Semantic Similarity} The semantic similarity of the best paraphrase and gold references, measured by the BERT and SBERT metrics, was in most cases higher for the supervised, trained GPT-2 and T5 models, and for the GPT-2 model trained using our self-supervised SMCLM method. We achieved results comparable with the supervised and reference ChatGPT methods on semantic similarity. We explain BackTransl's BERT and SBERT results as the syntactic similarity of the generated candidates to the source sentences. Our hypothesis is supported by the results of both lexical similarity to the source sentence and semantic similarity results, with consideration for lexical diversity. In terms of \textit{ori}BERT and \textit{ori}SBERT scores, the unsupervised methods, including SMCLM, also achieved high scores, which indicates high semantic similarity between the source sentences and the generated ones. We noted that the sensitivity of the SBERT measure was higher than that of the BERT measure, whose range of values returned was narrower.

\subsection{Lexically-dependent semantic similarity}
In terms of lexically-dependent semantic similarity, the supervised trained models, particularly GPT-2 and T5, achieved better results. Notably, the SMCLM methods achieved the best results for lexically-dependent metrics in unsupervised methods and comparable results with supervised methods for BERT-\textit{i}BLEU and SBERT-\textit{i}BLEU. These results indicate high semantic similarity of the generated paraphrases for these methods when deployed simultaneously with words other than those in the source sentence. This aligns with our expectations. We also observed better results for the SMCLM models trained using the 10M Sentence Corpus. The results of ChatGPT confirm the reliability of these metrics.  The values of lexically-dependent semantic similarity did not coincide with those of the BERT and SBERT measures. These results demonstrate that measures based solely on simple semantic similarity promote the generation of paraphrases in such a way that a large part of the sentence is copied from the source text. We consider this an undesirable effect. We recommend using measures that simultaneously consider semantic similarity and lexical diversity, such as BERT-\textit{i}BLEU and SBERT-\textit{i}BLEU. The lack of correlation between the measures also confirms our earlier hypothesis: that the reliability of the BLEU and ROUGE measures is low.

GPT-3.5 and GPT-4o-mini reference methods demonstrate outstanding fluency and strong semantic fidelity, while maintaining moderate lexical diversity. Notably, GPT-4o-mini achieves state-of-the-art results in BERT-iBLEU and SBERT-iBLEU, indicating that it generates paraphrases that are both semantically faithful and lexically nuanced—outperforming all other evaluated systems in this regard.

The average results across the three evaluated datasets (Table \ref{tab:res}, subtable \textit{Total}) reveal notable discrepancies between traditional semantic similarity metrics such as BLEU-3 and ROUGE-L, and metrics like BERT-\textit{i}BLEU and SBERT-\textit{i}BLEU, which incorporate inverse lexical similarity. Among the supervised methods, the T5 v1.1 model achieved the highest average scores across all lexical and semantic similarity measures. However, the GPT-2 model outperformed T5 v1.1 on BERT-\textit{i}BLEU and SBERT-\textit{i}BLEU metrics. This suggests that T5 v1.1 tends to generate paraphrases that are closer to the input—grammatically correct but lexically similar—thus yielding higher scores on metrics that do not penalize lexical overlap. In contrast, GPT-2 generates paraphrases that are both semantically accurate and lexically diverse, as reflected in its superior performance on BERT-\textit{i}BLEU, SBERT-\textit{i}BLEU, and \textit{ori}BLEU. These findings highlight the importance of incorporating metrics sensitive to lexical variation, such as BERT-\textit{i}BLEU and SBERT-\textit{i}BLEU, in the evaluation of paraphrase generation models.

A similar pattern is observed in the averaged results for the unsupervised methods. The BackTransl method achieves strong performance across most metrics; however, when lexical diversity with respect to the input is taken into account, the SMCLM models yield the highest scores on BERT-\textit{i}BLEU and SBERT-\textit{i}BLEU, achieving 83.44 and 82.73 respectively. This further underscores the effectiveness of SMCLM in generating semantically faithful yet lexically diverse paraphrases.

To provide a clearer picture of the generated paraphrases, tables \ref{tab:examples_smclm} and \ref{tab:examples_refs} includes representative examples for all evaluated methods.

\section{Limitations}
\label{sec:limitations}
Like the methods based on neural sequential models, our approach must address the low probability problem for rare and out-of-vocabulary tokens. We observed modifications in proper names and numbers in the generated paraphrases relative to the source sentences. Integrating with the Copy Mechanism \cite{gu-etal-2016-incorporating} or using semantic encoders that consider that kind of token might counter this effect. 

Another limitation of our method is the width of the context that the semantic embeddings consider. The generation of paraphrases for longer passages of text requires the use of embeddings that consider a wider context. This could affect the performance of our method.

Our research is concerned only with the English language. The availability of multilingual encoders may lead to further research on multilingual paraphrasing or translation with simultaneous paraphrasing. This area remains a promising direction for future research. Appendix \ref{sec:multilingual} presents examples of paraphrases generated by the model trained with the SMCLM method and the multilingual sentence transformer. 

\section{Ethical Considerations}
\label{sec:ethical_considerations}
All authors of this article acknowledge the ACM Code of Ethics and honor the code of conduct.

\paragraph{Data and Models.} No personal information was used or disclosed during our study. The QQP, MSCOCO, and CNN News datasets, as well as sentence encoders we used, are publicly available and used widely. All datasets used in our study came from Hugging Face Hub and comply with Hugging Face Hub's privacy policy. 

\paragraph{Source Code.}  We used only source code that has been released publicly by its authors in the reproduction of the paraphrasing models and the implementation of our solution. The code is available publicly in GitHub repositories.

\paragraph{Potential Risks.} We assume that our model might be used to generate plagiarized material, such as scientific articles, book content, and other copyrighted written content.

\section{Conclusion}
\label{sec:conclusions}
This article presents semantically meaningful causal language modeling (SMCLM), a novel self-supervised text generation method designed to leverage text encoders for semantically similar text generation. Extensive experiments on the paraphrase generation task demonstrate that SMCLM performs better than state-of-the-art unsupervised methods and is comparable to the supervised approaches of automatic paraphrase generation. This article also presents a broad set of metrics for comprehensive paraphrase generation evaluation. Our evaluation includes lexically dependent semantic similarity metrics, which we propose as a new standard for APG evaluation. Our study also proves that widely used metrics like BLEU, ROUGE, and BERTScore neglect to consider essential aspects of paraphrase evaluation.

\paragraph{Future work}
To mitigate issues related to rare words, proper names, and numerical expressions being incorrectly paraphrased, future work could explore the integration of mechanisms such as the Copy Mechanism or token-aware semantic encoders. These enhancements may improve fidelity in paraphrasing named entities and domain-specific terms.

Since current semantic embeddings may have limited context windows, future research could focus on extending SMCLM to handle long-form text, potentially through the use of hierarchical or document-level embeddings, retrieval-augmented encoders, or long-context transformers (e.g., Longformer, BigBird).

While the current model is English-only, there is strong potential to generalize the approach to multilingual settings, especially with the availability of multilingual sentence transformers like LaBSE, mBERT or BGE-M3. Future work could investigate multilingual paraphrasing, cross-lingual paraphrase generation, or even joint paraphrase-translation frameworks.

SMCLM has been evaluated primarily on structured or short-form data. A promising research avenue is to assess its generalizability to varied domains, such as legal, biomedical, or conversational text, and fine-tune or adapt SMCLM to these genres.

Finally, a key direction for future work is the application of SMCLM to larger-scale autoregressive models such as LLaMA or GPT-3. These more expressive architectures may significantly enhance both the quality and diversity of generated paraphrases, particularly in low-resource or long-context scenarios. Leveraging such models could unlock further performance gains and improve scalability across domains.

\appendices

\section{SMCLM-GPT-2 Implementation Details}
\label{sec:smclmgpt2}

\begin{figure}[h]
\centering
\includegraphics[height=6.7cm, width=8cm]{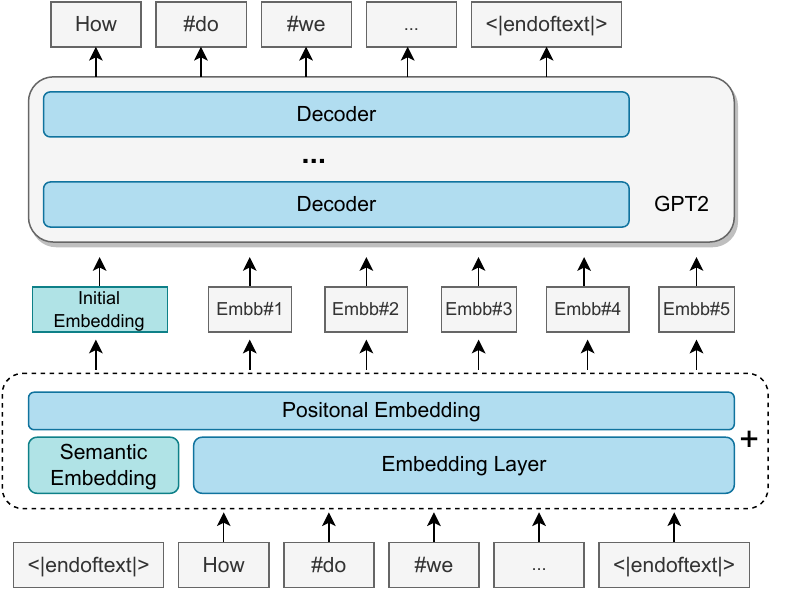}
\caption{The scheme of GPT2-based implementation of the SMCLM method.}
\label{fig:gpt2_scheme}
\end{figure}

Figure \ref{fig:gpt2_scheme} shows an implementation scheme of our GPT-2 model-based SMCLM approach. The semantic embedding of the sentence, encoded by the SentenceTransformers \textbf{paraphrase-mpnet-base-v2} encoder, is passed to the GPT-2 model. It replaces the first token's embedding from the GPT-2 embedding layer - in the case of our model, the token <$\vert$endoftext$\vert$>, which is added during tokenization. The size of the sentence encoding embedding must match the size of the token embedding in the model. In the case of our SMCLM implementation, both the GPT-2 model embedding layer and the \textbf{paraphrase-mpnet-base-v2} encoder map the input to a 768-dimensional dense vector space. The embedding is then summed with positional embedding, as it is in the original GPT-2 model. This approach ensures that the injected semantic vector is treated first in the token sequence. The source code of our implementation is publicly available\footnote{\url{https://github.com/mmichall/bleu-macaw}}.

\section{Hyperparameters and Setup}
\label{app:params}

\paragraph{Hyperparameters.} Table \ref{tab:hp} shows the hyperparameters used in our experiments for the SMCLM, GPT-2, and T5 v1.1 models. For the other methods, we used the parameters recommended by the authors and described in their publications. For the \textit{diverse beam search} decoding algorithm, we experimented with the following parameters: for \textit{diverse penalty} we tested values 0.2, 0.4, 0.6, 0.8; for \textit{beams number} we tested the values 5 and 10, and for \textit{groups number} we also tested the values 5 and 10; for \textit{repeat n-gram size} we set the value 2. Through experiments, we determined the best set of parameters for \textit{diverse beam search} as: \textit{diverse penalty} = 0.6; \textit{beams number} = 5; \textit{groups number} = 5; \textit{repeat n-gram size} = 2 for results and performance for SMCLM, GPT-2, and T5 v1.1 decoding methods. For other methods, we used the generation algorithms suggested in the publications.

\begin{table}[!ht]
\small
\setlength{\tabcolsep}{6pt} % Default value: 6pt
\renewcommand{\arraystretch}{1.2} % Default value: 1
\begin{tabular}{|l|lcc|}
\hline 
\textbf{Model} & \textbf{Parameter} & \textbf{Value} & \textbf{Tested values} \\
\hline &&& \\
\multirow{6}{*}{SMCLM} & learning rate & 5e-6 & 5e-7, 1e-6, 5e-6 \\
 & batch size & 32 & 32 \\
 & weight decay & 1e-2 & 1e-2, 1e-3 \\
 & optimizer & AdamW & AdamW \\
 & train epochs & 8 & 8 \\
 & warmup steps & 2000 & 2000 \\ &&&\\ \hline &&& \\

\multirow{5}{*}{GPT-2}  & learning rate & 1e-5 & 5e-6, 1e-5, 5e-5 \\
 & batch size & 16 & 16 \\
 & weight decay & 1e-2 & 1e-2, 1e-3 \\
 & optimizer & AdamW & AdamW \\
 & train epochs & 20 & 10, 20 \\
 & warmup steps & 2000 & 2000 \\ &&&\\ \hline &&& \\

\multirow{6}{*}{T5 v1.1}  & learning rate & 1e-6 & 5e-7, 1e-6, 5e-6 \\
 & batch size & 16 & 16 \\
 & weight decay & 1e-2 & 1e-2, 1e-3 \\
 & optimizer & AdamW & AdamW \\
 & train epochs & 10 & 10, 20 \\
 & warmup steps & 2000 & 2000 \\ &&&\\ \hline
\end{tabular}
\hspace{0.5em}

\caption{Hyperparameters used for SMCLM, GPT-2, and T5 v1.1 methods training, as well as ranges of parameters we experimented with.}
\label{tab:hp}
\end{table}

\paragraph{Setup}
We used NVIDIA Tesla V100 32GB during our experiment to train all models, including SMCLM ones. It takes around 60 hours to train the SMCLM-GPT-2 model with 124M parameters model on 10M Sentence Corpus for 8 epochs. For much smaller corpora created from datasets under study, it takes around 4, 8, and 12 hours to train or fine-tune a model for QQP, CNN News, and MSCOCO training splits, respectively, for 10 epochs.

\section{10M Sentence Corpus}
\label{app:10m}
To create the 10M Sentence Corpus, we select a set of datasets from Hugging Face Hub, considering the variety of domains from which the data comes. We select the following data domains: \textit{general}, \textit{stock}, \textit{news}, \textit{image captions}, \textit{wikipedia}, \textit{reviews}, \textit{questions}, \textit{prompts}, \textit{science}. Used datasets are listed in Table \ref{tab:10M_datasets}. To create the corpus containing 10M sentences, we use the pipeline as follows:
\begin{enumerate}
    \item Randomly choose one data domain of those listed above and randomly choose a dataset from the chosen domain.
    \item Then, randomly get an example text from the chosen dataset.
    \item Check the following conditions:
    \begin{enumerate}
         \item if the example text is shorter than 10 characters,
         \item if detected language of selected text example is not English (we use the fasttext-langdetect library\footnote{\url{https://pypi.org/project/fasttext-langdetect/}} for language detection).
    \end{enumerate}
    \item Get the next document from the dataset if any of the above conditions are met and go to the step 3. If not, go to the step 5. 
    \item Split the chosen text into sentences with NLTK sentence tokenizer\footnote{\url{https://www.nltk.org/api/nltk.tokenize.sent\_tokenize.html}} and randomly chose one sentence.
    \item Check if the sentence is unique by hashing it with built-in Python \textit{hash} method and check if the hash has not occurred before. If not, we add the sentence to the corpus. If so, get the next text example from the dataset and go to step 3.
\end{enumerate}

We repeated the pipeline until we achieved 10M sentences in our corpus.

\begin{table}[h]
\small
\setlength{\tabcolsep}{6pt} % Default value: 6pt
\renewcommand{\arraystretch}{1.2} % Default value: 1
\begin{tabular}{|l|l|}
\hline
\textbf{Dataset name} & \textbf{Category} \\
\hline 
vietgpt/the\_pile\_openwebtext2 & general \\ \hline
c4 & general \\ \hline
mjw/stock\_market\_tweets & stock \\ \hline
lvwerra/stack-exchange-paired & stock \\ \hline
ag\_news &  news \\ \hline
rjac/all-the-news-2-1-Component-one & news \\ \hline
conceptual\_captions & image captions\\ \hline
recastai/flickr30k-augmented-caption & image captions \\ \hline
Andyrasika/image\_captioning & image captions\\ \hline
squad\_v2 & questions\\ \hline
hotpot\_qa & questions\\ \hline
one-million-reddit-questions & questions\\ \hline
lucadiliello/newsqa & questions\\ \hline
maximedb/natural\_questions & questions\\ \hline
b-yukky/msmarco-short & questions \\ \hline
yahoo\_answers\_qa & questions\\ \hline
gooaq & questions \\ \hline
9wimu9/eli5\_mult\_answers\_en & questions \\ \hline
BeIR/nq & questions \\ \hline
Gustavosta/Stable-Diffusion-Prompts & prompts\\ \hline
ywchoi/pubmed\_abstract\_0 & science\\ \hline
ywchoi/pubmed\_abstract\_1 & science\\ \hline
arxiv\_dataset & science\\ \hline
\makecell[l]{wiki-en-passages-20210101} & wikipedia \\ \hline
polinaeterna/amazon\_us\_reviews & reviews\\ \hline

\end{tabular}
\caption{The Hugging Face Hub datasets with data categories used to build the 10M Sentence Corpus.}
\label{tab:10M_datasets}
\end{table}

\section{Performance and Memory Usage Evaluation of SMCLM Models}
\label{sec:d}

To better understand the computational requirements and efficiency of the SMCLM models, we conducted a comparative analysis of training time, memory consumption, and inference throughput. Table \ref{tab:performance} presents an overview of these metrics for selected SMCLM models variants of GPT-2 and T5 models. As shown, smaller models such as GPT-2 (124M) offer significantly faster training and higher inference throughput, making them suitable for resource-constrained environments. In contrast, larger models like T5-large, while more demanding in terms of memory and training time, can provide improved performance on complex downstream tasks. All experiments were conducted using the paraphrase-mpnet-base-v2 encoder on a single Tesla V100 32 GB GPU. Trainings was conducted using a batch size of 8 and the Adam optimizer with a weight decay of 0.001. A warmup ratio of 6\% of the total training steps was applied, and all models were trained for 3 epochs on 10K training examples limited to 512 tokens.

\begin{table}[!ht]
\renewcommand{\arraystretch}{1.2}
\begin{tabular}{lcccc}
\toprule
Model       & Size & \makecell{Training\\ Time} & \makecell{Training \\Memory\\ Used} & \makecell{Inference \\examples \\per second} \\ \hline
gpt2        & 124M &  34 min  & 6250 MB &   89   \\
gpt2-medium & 380M &  58 min  &  11400 MB  &  41  \\
gpt2-large  & 812M &  12 min  &  24600 MB  &  27 \\
t5-base     & 223M &  48 min  &  10200 MB  &  47 \\
t5-large    & 735M &  120 min &  24400 MB  &  29  \\
\bottomrule
\end{tabular}
\caption{Comparison of SMCLM language models based on gpt2, gpt2-medium, gpt2-large, t5-base, t5-large in terms of model size (number of parameters), training time, training memory usage, and inference throughput (examples per second). The results highlight trade-offs between model size, computational resource requirements, and inference efficiency.}
\label{tab:performance}
\end{table}

\section{Selection of the $\beta$ coefficient in the BERT-iBLEU and SBERT-iBLEU equations}
\label{sec:e}

The selection of the coefficient $\beta$ in the formulas for the BERT-\textit{i}BLEU and SBERT-\textit{i}BLEU metrics plays a crucial role in balancing the contribution of the two components involved—semantic similarity, quantified by the BERT or SBERT score (for BERT-\textit{i}BLEU and SBERT-\textit{i}BLEU, respectively), and inverse lexical similarity, measured using the \textit{i}BLEU score between the input sentence and the best generated paraphrase. Increasing the value of $\beta$ results in a stronger emphasis on the BERT/SBERT component, thus prioritizing semantic similarity.

In our study, we aimed to establish a balanced contribution of both components. To this end, we computed the average values of the BERT, SBERT, and \textit{i}BLEU scores across all examples in the QQP, MSCOCO, and CNN News datasets. To ensure an unbiased evaluation of the automatic paraphrasing methods and to avoid tuning the $\beta$ parameter to favor any particular model's output, we used reference paraphrase pairs (input–reference paraphrase) drawn from the respective datasets as the basis for our calculations. A detailed description of these reference pairs is provided in Section \ref{sec:datasets}, specifically in the subsection describing the construction of supervised datasets.

Based on the ratio of the BERT, SBERT, and \textit{i}BLEU values, we set the value of the $\beta$ parameter to 2 in our formulas to balance semantic and lexical similarity in the BERT-\textit{i}BLEU and SBERT-\textit{i}BLEU metrics. The averaged values of the evaluated parameters are presented in Table \ref{tab:ratio}.

Table \ref{tab:beta} reports the BERT-\textit{i}BLEU and SBERT-\textit{i}BLEU scores for the SMCLM-10M-ft model, averaged across the three datasets—QQP, MSCOCO, and CNN News—as a function of the $\beta$ parameter, which was varied over integer values from 1 to 5.

% \usepackage{multirow}
\begin{table}[]
\renewcommand{\arraystretch}{1.2}
\label{tab:ratio}
\begin{tabular}{ccccc}
\toprule
\multicolumn{1}{c}{\multirow{2}{*}{BERT}} & \multicolumn{1}{c}{\multirow{2}{*}{SBERT}} & \multicolumn{1}{c}{\multirow{2}{*}{\textit{i}BLEU}} & \multicolumn{2}{c}{ratio}                        \\ \cline{4-5} 
\multicolumn{1}{c}{}                      & \multicolumn{1}{c}{}                       & \multicolumn{1}{c}{}                       & \multicolumn{1}{c}{BERT / \textit{i}BLEU} & SBERT / \textit{i}BLEU \\ \hline
                82.39                     &              78.49                         &              40.9                        &               2,01                   &       1.92        \\
\bottomrule
\end{tabular}
\caption{Average values of the BERT, SBERT, and BLEU scores along with the BERT/\textit{i}BLEU and SBERT/\textit{i}BLEU ratios for all input–reference paraphrase pairs from the QQP, MSCOCO, and CNN News datasets.}
\label{tab:ratio}
\end{table}

\begin{table}[!ht]
\centering
\renewcommand{\arraystretch}{1.2}
\begin{tabular}{lccccc}
\toprule
$\beta$ & 1 & 2 & 3 & 4 & 5 \\ \hline
BERT-\textit{i}BLEU & 85.57 & 83.44 & 82.44 & 81.84 & 81.44 \\ \hline
SBERT-\textit{i}BLEU & 83.87 & 82.73 & 82.17 & 81.83 & 81.61 \\
\bottomrule
\end{tabular}
\caption{Averaged BERT-\textit{i}BLEU and SBERT-\textit{i}BLEU scores averaged across the three evaluated datasets for the SMCLM-10M-ft model depending on the value of the $\beta$ parameter.}
\label{tab:beta}
\end{table}

\section{Multilingual SMCLM Model}
\label{sec:multilingual}

By integrating multilingual encoders into SMCLM models, the models acquire the capability to generate paraphrases of input sentences across multiple languages supported by the encoder. This cross-lingual paraphrasing ability opens up promising directions for further research, particularly in the context of multilingual understanding and generation. Table \ref{tab:multi} presents representative paraphrases produced by the SMCLM-GPT model trained with the \textit{paraphrase-multilingual-mpnet-base-v2} encoder, for input sentences originally written in German, French, and Spanish.

\begin{table*}[!th]
\small
\setlength{\tabcolsep}{5pt} % Default value: 6pt
\renewcommand{\arraystretch}{0.4} % Default value: 1
\begin{tabular}{ll}
\hline
\textbf{Source sentence} & \textbf{Best paraphrase} \\ \hline \\
\multicolumn{2}{c}{QQP} \\ \hline
\\
\makecell[l]{Which are best places to live in China?}  &
\makecell[l]{Which are the best places in China to live?} \\
\\

\makecell[l]{What are the characteristics of a person with \\ an INFP personality?}  &
\makecell[l]{What are personality traits of an INFP?} \\
\\
\makecell[l]{Clinton or Trump: who is a better option for \\ the USA?}  &
\makecell[l]{Which is a better choice for the US: Trump \\ or Clinton?} \\
\\

\makecell[l]{What are some great activities you can do at\\ a sleepover with your best friend?}  &
\makecell[l]{What are some awesome things to do with your\\ friend during your sleepover?} \\
\\

\makecell[l]{How do I get ripped in 3 months?}  &
\makecell[l]{How can I cut my body in 3 months?} \\
\\

\makecell[l]{Why Indian government does not take any control\\ measures for over population?}  &
\makecell[l]{Why the Indian Government has not control over \\population?} \\
\\	

\makecell[l]{What are the reasons behind nuclear energy being \\non-renewable?}  &
\makecell[l]{What is the main reason nuclear energy is not renewable?} \\
\\	

\makecell[l]{What's your New Year 2017 resolution?}  &
\makecell[l]{What is your new year's resolution?} \\
\\
\multicolumn{2}{c}{MSCOCO} \\ \hline
\\
\makecell[l]{A man has jumped a ramp with his skateboard.}  &
\makecell[l]{A man jumping a skateboard off a ramp.} \\
\\

\makecell[l]{Soccer players in orange and grey uniforms on a field.}  &
\makecell[l]{A soccer team wearing orange and yellow uniforms\\ are on the field.} \\
\\

\makecell[l]{There is a pizza with many toppings on it}  &
\makecell[l]{A pizza that has lots of toppings on it.} \\
\\
 		
\makecell[l]{three strange looking birds walking on the \\ grass}  &
\makecell[l]{Three birds walking in the grass.} \\
\\
\makecell[l]{A blue train is next to a sidewalk on the \\ rails.}  &
\makecell[l]{A blue train is parked on a sidewalk.} \\
\\
\makecell[l]{A train with people inside is about to go into \\ a tunnel.}  &
\makecell[l]{A train is in the train tunnel and people are in it.} \\
\\
\makecell[l]{A stop sign is propped up against the side of \\ a building.}  &
\makecell[l]{A stop sign has been placed upside-down in the \\ grass beside a building.} \\
\\
\makecell[l]{A cat and dog napping together on the couch.}  &
\makecell[l]{A cat and a dog laying on their couch.} \\
\\
\multicolumn{2}{c}{CNN News} \\ \hline
\\
\makecell[l]{She claimed she was sorry, though continued\\ to say she killed in self-defence.}  &
\makecell[l]{She stated that she was sorry for her \\self-defense and that it was her duty to do so.} \\
\\
\makecell[l]{And he's sharing the pain, cutting his own office \\budget by a like amount.}  &
\makecell[l]{He's cutting his own share of the budget.} \\
\\
\makecell[l]{She said she had 20,000 euros 'around me'.}  &
\makecell[l]{She said that she had 20,000 euros in her pocket.} \\
\\
\makecell[l]{which has been comforting in what has been a\\ horrendous two weeks.}  &
\makecell[l]{It's been a relief for two weeks that have been\\ somewhat unappealing.} \\
\\
\makecell[l]{The reason is simple: With a dearth of\\ infrastructure, the vast majority of people (an \\ estimated 1.5-billion globally, according\\ to the UN) have no electricity.}  &
\makecell[l]{According to the UN, about 1.5 billion people\\ have no access to electricity, and that\\ is why the majority of the population is in\\ a state of low-power.} \\
\\
\makecell[l]{More than 60,000 people, many of them children, \\have fled for safer areas in Pakistan and\\ Afghanistan, according to a United \\Nations-supported news network that reports on\\ humanitarian efforts.}  &
\makecell[l]{More than 60,000 people have fled their homes in\\ the past year, according to a U.N. report, citing\\ officials.} \\
\\
\makecell[l]{The program's code is also written to evolve \\over time and its author appears to be making updates to\\ thwart attempts to neuter the worm.}  &
\makecell[l]{The program is being updated to maintain the \\worm's ability to evolve.} \\
\\
\makecell[l]{ISIS' influence has spread cross the Middle East and\\ north Africa, with groups of terrorists in Libya and \\Egypt as well as the terror network's strongholds of\\ Iraq and Syria} &
\makecell[l]{ISIS has spread across the region, with the \\group's influence in Iraq and Syria.} \\

\end{tabular}
\caption{Examples of paraphrases generated by SMCLM-GPT-2 models pre-trained on the 10M Sentence Corpus and fine-tuned on data derived from datasets.}
\label{tab:examples_smclm}
\end{table*}

\begin{table*}[!th]
\small
\setlength{\tabcolsep}{5pt} % Default value: 6pt
\renewcommand{\arraystretch}{0.4} % Default value: 1
\begin{tabular}{ll}

Original sentence & GPT 3.5\\ \hline
\\
\makecell[l]{Which are best places to live in China?}  &
\makecell[l]{What are the top cities to live in within China?}  \\
\\
\makecell[l]{What are the characteristics of a person with \\ an INFP personality?}  &
\makecell[l]{What traits define someone with an INFP personality type?}   \\
\\
\makecell[l]{Clinton or Trump: who is a better option for \\ the USA?}  &
\makecell[l]{Between Clinton and Trump, who would be a better\\ choice for the United States?}   \\
\\
\makecell[l]{A man has jumped a ramp with his skateboard.}  &
\makecell[l]{A man performed a jump off a ramp using his skateboard.}   \\
\\
\makecell[l]{Soccer players in orange and grey uniforms on a field.}  &
\makecell[l]{Athletes in orange and grey jerseys playing soccer on the field.}   \\
\\
\makecell[l]{There is a pizza with many toppings on it}  &
\makecell[l]{A pizza loaded with a variety of toppings is on display.}   \\
\\
\makecell[l]{She claimed she was sorry, though continued\\ to say she killed in self-defence.}  &
\makecell[l]{She insisted she was sorry, but maintained it was \\an act of self-defense.}  \\
\\
\makecell[l]{And he's sharing the pain, cutting his own office \\budget by a like amount.}  &
\makecell[l]{He’s also feeling the impact, trimming his own \\office budget by a similar amount.}   \\
\\
\makecell[l]{She said she had 20,000 euros 'around me'.}  &
\makecell[l]{She said she had 20,000 euros on her.}   \\
\\
\makecell[l]{}  &
\makecell[l]{} \\
\\

GPT-4o-mini & DiPS \\ \hline
\\

\makecell[l]{Which locations in China are considered the best for living?} &
\makecell[l]{What are the best cities to live in China?} \\
\\
\makecell[l]{How would you describe a person who is an INFP?} &
\makecell[l]{What are the characteristics of a person with \\ an dreams?} \\
\\
\makecell[l]{Who is the preferable candidate for America: Clinton or Trump?} &
\makecell[l]{Who will be the next president of the United States?} \\
\\
\makecell[l]{A skateboarder has launched himself over a ramp.} &
\makecell[l]{A man doing skateboard tricks in front of a crowd.} \\
\\
\makecell[l]{Soccer players wearing orange and grey kits are on the pitch.} &
\makecell[l]{A group of soccer players on a soccer field.} \\
\\
\makecell[l]{The pizza is covered with numerous different toppings.} &
\makecell[l]{A close up of a sliced pizza on a plate} \\
\\
\makecell[l]{Although she expressed remorse, she continued to\\ assert that the killing was in self-defense.} &
\makecell[l]{Despite acknowledging her innocence in self defense, \\ she maintained that she was a pity} \\
\\
\makecell[l]{To share the burden, he’s reducing his office\\ budget by the same figure.} &
\makecell[l]{He's fixated on the pain and UNK of his own office.} \\
\\
\makecell[l]{According to her, she was carrying about \\20,000 euros.} &
\makecell[l]{She stated that she had connected with 20,000 euros.} \\
\\
\makecell[l]{}  &
\makecell[l]{} \\

QCPG & CGMH \\ \hline
\\
\makecell[l]{Which are the best places to visit in China?}  &
\makecell[l]{Which are the best places to visit in China?} \\
\\
\makecell[l]{What are the characteristics of a person with \\ an PPP character?}  &
\makecell[l]{What are the qualities of a person with \\ an INFP personality?} \\
\\
\makecell[l]{Clinton or Trump: who is a better option for \\ the president?}  &
\makecell[l]{Clinton or Spotify: who is a better option for \\ the USA?} \\
\\
\makecell[l]{A man has jumped a off with his skate.}  &
\makecell[l]{A man that just jumped off a ramp with his skateboard.} \\
\\
\makecell[l]{Soccer players in yellow and grey uniforms on the field.}  &
\makecell[l]{Baseball players in yellow and grey uniforms are on a field.} \\
\\
\makecell[l]{There is the pizza with lot toppings on}  &
\makecell[l]{There is a very large pizza with some toppings on it} \\
\\
\makecell[l]{She claimed she was sorry, though continued\\ to say she killed in self-defence.}  &
\makecell[l]{She claimed she was sorry, though continued\\ to say she killed in UNK} \\
\\
\makecell[l]{And he's sharing the pain, cutting his own office \\money by a like amount.}  &
\makecell[l]{If he's sharing the pain, cutting his own office budget by a} \\
\\
\makecell[l]{She said she has 20,000 euros 'around me'.}  &
\makecell[l]{He said that had 20,000 euros UNK me.} \\
\\
\makecell[l]{}  &
\makecell[l]{} \\
\end{tabular}
\end{table*}

\begin{table*}[!th]
\small
\setlength{\tabcolsep}{5pt} % Default value: 6pt
\renewcommand{\arraystretch}{0.4} % Default value: 1
\begin{tabular}{ll}

T5 v1.1 & GPT-2 \\ \hline
\\
\makecell[l]{In China, which is best places to live?}  &
\makecell[l]{Which are the best cities to live in China?} \\
\\
\makecell[l]{What are the characteristics of a person with\\  an INFP personality ?}  &
\makecell[l]{What are the characteristics of a person who \\ has an INFP personality?} \\
\\
\makecell[l]{Clinton or Trump: Who is a better option for \\ the United States? }  &
\makecell[l]{Clinton or Trump: who is a better option for\\  the United States?} \\
\\
\makecell[l]{One man with his skate jumps a bit.}  &
\makecell[l]{A man jumped with his skate from the place I\\ am falling.} \\
\\
\makecell[l]{Football players in orange and grey uniforms on \\a field.}  &
\makecell[l]{Soccer players in orange and gray uniforms on a \\field.} \\
\\
\makecell[l]{There is a pizza with many toppings on it.}  &
\makecell[l]{Es has a pizza with many toppings on it.} \\
\\
\makecell[l]{She maintained that she was sorry, although continued\\ to say she killed herself in self-defence.}  &
\makecell[l]{And he takes the same pain, cutting his own  budget\\ in his own office in a similar amount} \\
\\
\makecell[l]{And he is sharing the pain by cutting his own office \\ budget by an amount like that.}  &
\makecell[l]{What are the reasons behind nuclear energy  being \\non-renewable?} \\
\\
\makecell[l]{She says she had 20,000 euros  around me''.}  &
\makecell[l]{That she had 20,000 euros 'round she.'} \\

\\
\makecell[l]{}  &
\makecell[l]{} \\
BackTransl & CorruptLM \\ \hline
\\

\makecell[l]{What are the best places to live in China?}  &
\makecell[l]{Which are the best cities to live in China?} \\
\\
\makecell[l]{What are the characteristics of a person with an INFP personality?}  &
\makecell[l]{What are the characteristics of a person who an INFP personality?} \\
\\
\makecell[l]{Clinton or Trump: Who's a better option for the United States?}  &
\makecell[l]{Clinton or Trump: who is a better candidate for the United States?} \\
\\
\makecell[l]{A man has jumped the ramp with his skateboard.}  &
\makecell[l]{A man jumped a ramp with his skateboard.} \\
\\

\makecell[l]{The football players in orange and grey uniforms on a field.}  &
\makecell[l]{The football players in orange and grey uniforms on the field.} \\
\\
\makecell[l]{There's a pizza with lots of toppings on it.}  &
\makecell[l]{There is the pizza with many toppings on it.} \\
\\
\makecell[l]{She said she was sorry, although she continued to say that \\she had killed in self-defence.}  &
\makecell[l]{She claimed she was sorry, although she continued to say that \\she had killed in self-defence.} \\
\\
\makecell[l]{And he shares the pain, cutting his own office budget from \\a similar amount.}  &
\makecell[l]{And he's sharing the pain, cutting his own office \\budget from \\a similar amount.} \\
\\
\makecell[l]{She said she had 20,000 euros around me.}  &
\makecell[l]{She said she has 20,000 euros "around me".} \\
\\

\end{tabular}
\caption{Examples of paraphrases generated by ChatGPT 3.5, ChatGPT-4o-mini, DiPS, QCPG, CGMH, T5 v1.1, GPT-2, BackTransl, CorrputLM models.}
\label{tab:examples_refs}
\end{table*}

\begin{table*}[!th]
\small
\setlength{\tabcolsep}{5pt} % Default value: 6pt
\renewcommand{\arraystretch}{0.4} % Default value: 1
\begin{tabular}{ll}
\hline
\textbf{Source sentence} & \textbf{Best paraphrase} \\ \hline \\
\multicolumn{2}{c}{German} \\ \hline
\\
\makecell[l]{Im Oktober 1560 traf er sich in Paris heimlich mit dem \\englischen Botschafter Nicolas Throckmorton und bat ihn um \\einen Pass, um durch Schottland nach England zurückzukehren.}  &
\makecell[l]{In October 1560 he secretly met with the English ambassador\\ to Paris, Nicolas Throckmorton, asking him for a passport\\ to return to England via Scotland.
} \\
\\
\makecell[l]{Die NBA-Saison 1975 - 76 war die 30. Saison der National\\ Basketball Association.}  &
\makecell[l]{The NBA season 1975 -- 76 was the 30th season of the \\National Basketball Association.} \\
\\
\makecell[l]{Es gibt auch spezifische Diskussionen, öffentliche \\Profildiskussionen und Projektdiskussionen.}  &
\makecell[l]{There are also specific discussions, public profile \\discussions, and project debates.} \\
\\

\makecell[l]{}  &
\makecell[l]{} \\
\multicolumn{2}{c}{French} \\ \hline
\\

\makecell[l]{À Paris, en octobre 1560, il rencontra secrètement l'ambassadeur \\d'Angleterre, Nicolas Throckmorton, lui demandant un passeport\\ pour retourner en Angleterre en passant par l'Écosse.}  &
\makecell[l]{In Paris, he met in October 1560 the English ambassador,\\ Nicholas Throckmorton, to demand him a passport for entry to\\ England via Scotland.
} \\
\\
\makecell[l]{La saison NBA 1975 - 76 était la 30e saison de la National \\Basketball Association.}  &
\makecell[l]{The NBA season of 1975 -- 76 was the 30th season of the National \\Basketball Association.} \\
\\
\makecell[l]{Il y a aussi des discussions spécifiques, des débats publics et des \\discussions de projet.}  &
\makecell[l]{There are also specific discussions, comments and projects.} \\
\\

\makecell[l]{}  &
\makecell[l]{} \\
\multicolumn{2}{c}{Spanish} \\ \hline
\\

\makecell[l]{En París, en octubre de 1560, se reunió en secreto con el embajador\\ inglés, Nicolas Throckmorton, pidiéndole un pasaporte para regresar\\ a Inglaterra a través de Escocia.}  &
\makecell[l]{In October 1560, he secretly met the English\\ ambassador, Nicolas Throckmorton in Paris, to ask for a passport \\to return through Scotland to England.} \\
\\
\makecell[l]{La temporada de la NBA de 1975: 76 fue la 30ª temporada de la \\National Basketball Association.}  &
\makecell[l]{The 1975 -- 76 NBA season was the 30th season of the \\National Basketball Association.} \\
\\
\makecell[l]{También hay discusiones específicas, debates de perfil público y \\discusiones de proyectos.}  &
\makecell[l]{There are also specific discussions, public profile debates \\and project discussions.} \\
\\

\end{tabular}
\caption{Examples of paraphrases generated by SMCLM model trained with multilingual \textit{paraphrase-multilingual-mpnet-base-v2} for examples in German, French and English.}
\label{tab:multi}
\end{table*}

\clearpage
\bibliography{acl2020}

\begin{IEEEbiography}[{\includegraphics[width=1.01in,height=1.25in,clip,keepaspectratio]{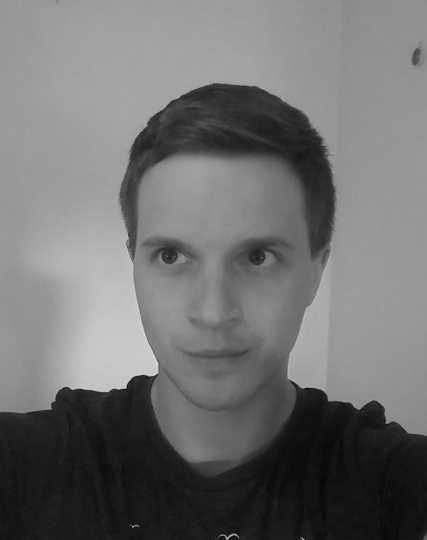}}]{MichaŁ PereŁkiewicz} received the M.S. degree in computer science from Lodz University of Technology, Lodz, Poland, in 2013. Currently works as a Principal Research and Technical Specialist in the AI Lab at the National Information Processing Institute in Warsaw, Poland. For several years professionally engaged in implementation of information systems incorporating machine learning solutions, primarily in the field of natural language processing.
His research interests focus on natural language processing issues including large language models, sentiment analysis, text embeddings and automatic paraphrasing.
\end{IEEEbiography}

\begin{IEEEbiography}[{\includegraphics[width=1in,height=1.25in,clip,keepaspectratio]{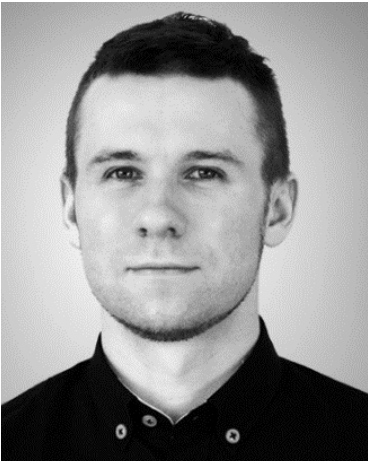}}]{SŁawomir Dadas}  received the Ph.D. degree in computer science with the Systems Research Institute, Polish Academy of Sciences, Warsaw, Poland. 
Currently works as a Deputy Head of the AI Lab at the National Information Processing Institute. For several years professionally engaged in design and implementation of information systems incorporating machine learning solutions, primarily in the field of natural language processing. 
His research interests include natural language processing, applications of machine learning in scientometric research, distributed computing, algorithms, and data structures.
\end{IEEEbiography}
%If you do not have or do not want to include a photo, you can use IEEEbiographynophoto as shown below:

\begin{IEEEbiography}
[{\includegraphics[width=1in,height=1.25in,clip,keepaspectratio]{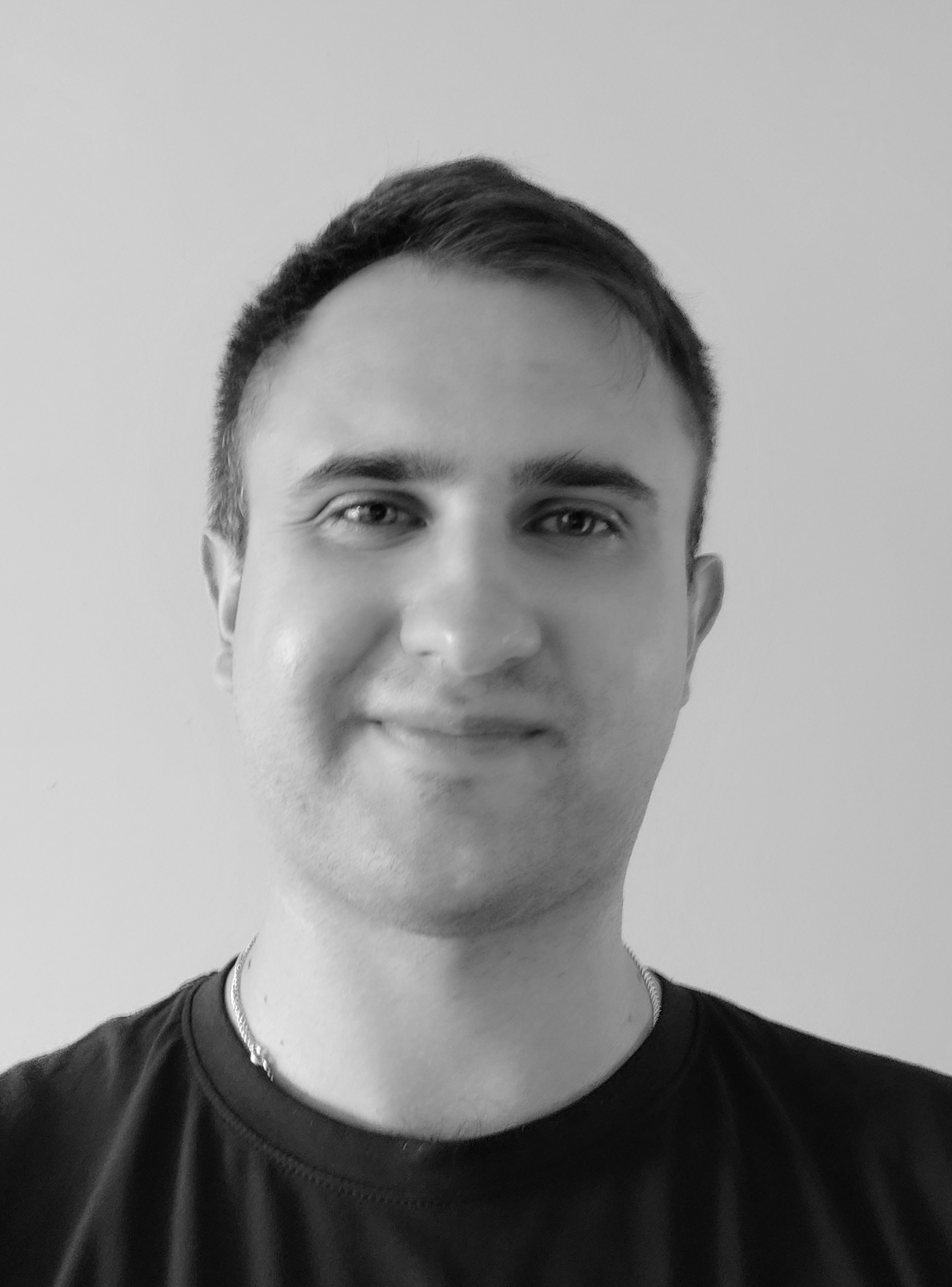}}]{RafaŁ PoŚwiata} holds an M.S. degree in electronic engineering from Military University of Technology in Warsaw, Poland, which he obtained in 2013. Since 2014, he has been working at the National Information Processing Institute, where he currently holds the position of Principal Research and Technical Specialist. His professional work primarily involves the development of information systems that incorporate machine learning solutions, with a particular focus on natural language processing.
His research interests include natural language processing, in particular topics such as emotion recognition, sentiment analysis, text embeddings and evaluation of large language models.
\end{IEEEbiography}

\EOD

\end{document}